\documentclass[10pt,twocolumn,letterpaper]{article}

\usepackage{cvpr}
\usepackage{times}
\usepackage{epsfig}
\usepackage{graphicx}
\usepackage{amsmath}
\usepackage{amssymb}
\usepackage{multirow}
\usepackage{booktabs}
\usepackage{pdfpages}

\usepackage[breaklinks=true,colorlinks=true,pagebackref=true,bookmarks=false]{hyperref}

\cvprfinalcopy 

\def\cvprPaperID{6640} 

\DeclareMathOperator*{\expectation}{\mathop{\mathbb{E}}}

\ifcvprfinal\fi
\begin{document}

\title{{Towards Inheritable Models for Open-Set Domain Adaptation}}

\author{Jogendra Nath Kundu\thanks{Equal Contribution} \quad Naveen Venkat\footnotemark[1] \quad Ambareesh Revanur \quad Rahul M V \quad R. Venkatesh Babu\\
Video Analytics Lab, CDS, Indian Institute of Science, Bangalore\\
}

\maketitle
\thispagestyle{empty}

\begin{abstract}
There has been a tremendous progress in Domain Adaptation (DA) for visual recognition tasks. Particularly, open-set DA has gained considerable attention wherein the target domain contains additional unseen categories. Existing open-set DA approaches demand access to a labeled source dataset along with unlabeled target instances. However, this reliance on co-existing source and target data is highly impractical in scenarios where data-sharing is restricted due to its proprietary nature or privacy concerns. Addressing this, we introduce a practical DA paradigm where a source-trained model is used to facilitate adaptation in the absence of the source dataset in future. To this end, we formalize knowledge inheritability as a novel concept and propose a simple yet effective solution to realize inheritable models suitable for the above practical paradigm. Further, we present an objective way to quantify inheritability to enable the selection of the most suitable source model for a given target domain, even in the absence of the source data. We provide theoretical insights followed by a thorough empirical evaluation demonstrating state-of-the-art open-set domain adaptation performance. Our code is available at {\small \tt \textbf{https://github.com/val-iisc/inheritune}}.

\end{abstract}


\vspace{-4.6mm}
\section{Introduction}
\vspace{-1.4mm}

Deep neural networks perform remarkably well when the training and the testing instances are drawn from the same distributions. However, they lack the capacity to generalize in the presence of a \textit{domain-shift} \cite{shimodaira2000improving} exhibiting alarming levels of dataset bias or domain bias \cite{datasetbias}. As a result, a drop in performance is observed at test time if the training data (acquired from a \textit{source} domain) is insufficient to reliably characterize the test environment (the \textit{target} domain).
This challenge arises in several Computer Vision tasks \cite{nath2018adadepth,long2015fully,kundu2019_um_adapt}
where 
one is often confined to a limited array of available source datasets, which are practically inadequate to represent a wide range of {target} domains.
This has motivated a line of Unsupervised Domain Adaptation (UDA) works that aim to generalize a model to an unlabeled {target} domain, in the presence of a labeled {source} domain.

In this work, we study UDA in the context of image recognition. 
Notably, a large body of UDA methods is inspired by the potential of deep CNN models to learn transferable representations \cite{howtransferable}. This has formed the basis of several UDA works that learn \textit{domain-agnostic} feature representations \cite{long2015learning,sun2016deepcoral,tzeng2014deep} by aligning the marginal distributions of the {source} and the {target} domains in the latent feature space. Several other works learn \textit{domain-specific} representations via independent domain transformations \cite{tzeng2017adversarial,dsbn,nath2018adadepth} to a common latent space on which the classifier is learned.
The latent space alignment of the two domains permits the reuse of the {source} classifier for the {target} domain. These methods however operate under the assumption of a shared label-set ($\mathcal{C}_s=\mathcal{C}_t$) between the two domains (\textit{closed-set}). This restricts their real-world applicability where a {target} domain often contains additional unseen categories beyond those found in the source domain.

\begin{figure}
    \label{vendor_client_paradigm}
    \centering
    \includegraphics[width=\columnwidth]{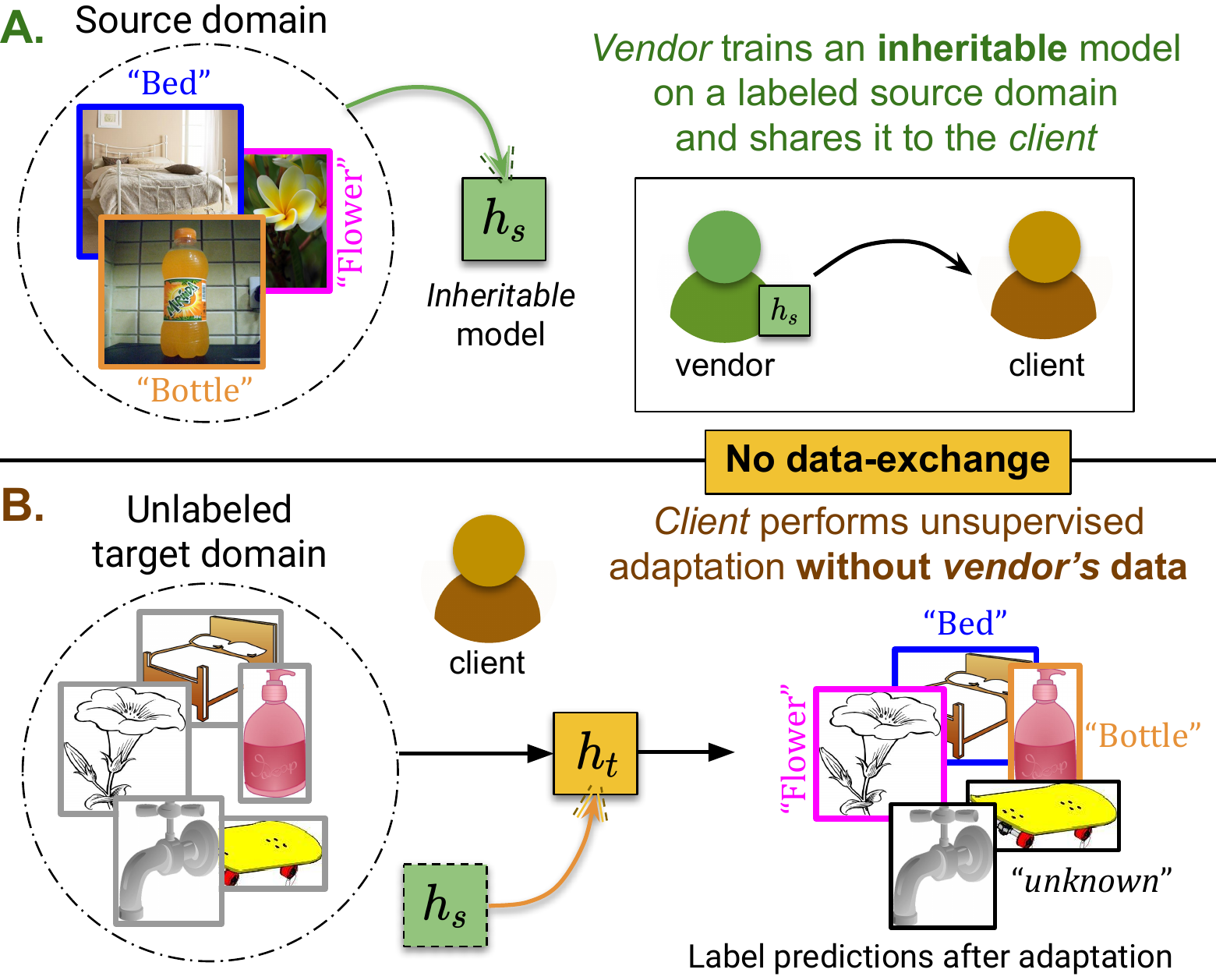}
    \caption{\small \textbf{A)} We propose \textit{inheritable} models to transfer the \textit{task-specific} knowledge from a model \textit{vendor} to the \textit{client} for, \textbf{B)} performing unsupervised \textit{open-set} domain adaptation in the absence of data-exchange between the \textit{vendor} and the \textit{client}. 
    }
    \vspace{-4mm}
    \label{fig:my_label}
\end{figure}



Recently, \textit{open-set} DA \cite{panareda2017open,saito2018open} has gained much attention, wherein the {target} domain is assumed to have unshared categories ($\mathcal{C}_s\subset\mathcal{C}_t$), a.k.a \textit{category-shift}. Target instances from the unshared categories are assigned a single \textit{unknown} label \cite{panareda2017open} (see Fig.~\ref{vendor_client_paradigm}{\color{red}B}). 
\textit{Open-set} DA is more challenging, since a direct application of distribution alignment (\eg as in \textit{closed-set} DA \cite{li2017mmd,sun2016deepcoral}) reduces the model's performance due to the interference from the unshared categories (an effect known as \textit{negative-transfer} \cite{pan2009survey}). 
The success of \textit{open-set} DA relies not only on the alignment of shared classes, but also on the ability to mitigate \textit{negative-transfer}. State-of-the-art methods such as \cite{UDA_2019_CVPR} 
train a domain discriminator using the source and the target data to
detect and reject {target} instances that are out of the {source} distribution, thereby minimizing the effect of \textit{negative-transfer}.

In summary, the existing UDA methods assume access to a labeled {source} dataset to obliquely receive a \textit{task-specific} supervision during adaptation.
However, this assumption of co-existing source and target datasets poses a significant constraint in the modern world, where coping up with strict digital privacy and copyright laws is of prime importance \cite{zskd}. This is becoming increasingly evident in modern corporate dealings, especially in the medical and biometric industries, where a source organization (the model \textit{vendor}) is often restricted to share its proprietary or sensitive data, alongside a pre-trained model to satisfy the \textit{client's} specific deployment requirements \cite{chidlovskii2016domain,hynes2018efficient}. Likewise, the \textit{client} is prohibited to share private data to the model \textit{vendor} \cite{federatedlearning}.
Certainly, the collection of existing \textit{open-set} DA solutions is inadequate to address such scenarios.

Thus, there is a strong motivation to develop practical UDA algorithms which make no assumption about data-exchange between the \textit{vendor} and the \textit{client}. One solution is to design self-adaptive models that effectively capture the \textit{task-specific} knowledge from the \textit{vendor's} source domain and transfer this knowledge to the \textit{client's} target domain.
We call such models as \textit{inheritable} models, referring to their ability to inherit and transfer 
knowledge across domains without accessing the source domain data.
It is also essential to quantify the knowledge \textit{inheritability} of such models. Given an array of \textit{inheritable} models, this quantification will allow a \textit{client} to flexibly choose the most suitable model for the \textit{client's} specific target domain.

Addressing these concerns, in this work we demonstrate how a \textit{vendor} can develop an \textit{inheritable} model, which can be effectively utilized by the \textit{client} to perform unsupervised adaptation to the {target} domain, without any data-exchange. To summarize, our prime contributions are:
\vspace{-1mm}
\begin{itemize}
    \setlength\itemsep{0mm}
    \item We propose a practical UDA scenario by relaxing the assumption of co-existing source and target domains, called as the \textit{vendor-client} paradigm.
    
    
    \item We propose \textit{inheritable} models to realize \textit{vendor-client} paradigm in practice and present an objective measure of \textit{inheritability}, which is crucial for model selection.
    
    
    \item We provide theoretical insights and extensive empirical evaluation to demonstrate state-of-the-art open-set DA performance using \textit{inheritable} models.

\end{itemize}{}

\section{Related Work}
\label{sec:related_work}


\noindent \textbf{Closed-set DA.} Assuming a shared label space ($\mathcal{C}_s=\mathcal{C}_t$), the central theme of these methods is to minimize the distribution discrepancy. Statistical measures such as MMD \cite{yan2017mindtheclassweightbias_weightedmmd,long2016unsupervised,long2017deepJAN}, CORAL \cite{sun2016deepcoral} and adversarial feature matching techniques \cite{ganin2016domain,tzeng2014deep,tzeng2015simultaneous,tzeng2017adversarial,Sankaranarayanan_2018_CVPR_GTA} are widely used. Recently, domain specific normalization techniques \cite{adabn,dsbn,cariucci2017autodial,featurewhiteningandconsensusloss} has started gaining attention. However, due to the shared label-set assumption these methods are highly prone to \textit{negative-transfer} in the presence of new target categories.


\vspace{2mm}

\noindent \textbf{Open-set DA.} ATI-$\lambda$ \cite{panareda2017open} assigns a pseudo class label, or an \textit{unknown} label, to each target instance based on its distance to each source cluster in the latent space. OSVM \cite{openSetjain2014multi} uses a class-wise confidence threshold to classify target instances into the source classes, or reject them as \textit{unknown}. OSBP \cite{saito2018open} and STA \cite{sta_open_set} align the source and target features through adversarial feature matching. However, both OSBP and ATI-$\lambda$ are hyperparameter sensitive and are prone to \textit{negative-transfer}. In contrast, STA \cite{sta_open_set} learns a separate network to obtain instance-level weights for target samples to avoid \textit{negative-transfer} and achieves state-of-the-art results. 
All these methods assume the co-existance of source and target data, while our method makes no such assumption and hence has a greater practical significance.

\vspace{2mm}

\noindent \textbf{Domain Generalization.} Methods such as \cite{d2018domain_DomGen,li2017deeper_DomGen,ding2017deep_DomGen,li2019episodic_DomGen,muandet2013domain_DomGen,khosla2012undoing_DomGen} largely rely on an arbitrary number of co-existing source domains with shared label sets, to generalize across unseen target domains. This renders them impractical when there is an inherent \textit{category-shift} among the data available with each \textit{vendor}. In contrast, we tackle the challenging \textit{open-set} scenario by learning on a single source domain.

\vspace{2mm}

\noindent \textbf{Data-free Knowledge Distillation (KD).} In a typical KD setup \cite{hinton2015distilling}, a student model is learned to match the teacher model's output. Recently, DFKD \cite{dfkd} and ZSKD \cite{zskd} demonstrated knowledge transfer to the student when the teacher's training data is not available. Our work is partly inspired by their \textit{data-free} ideology. However, our work differs from KD in two substantial ways; 1) by nature of the KD algorithm, it does not alleviate the problem of \textit{domain-shift}, since any domain bias exhibited by the teacher will be passed on to the student, and 2) KD can only be performed for the task which the teacher is trained on, and is not designed for recognizing new (\textit{unknown}) target categories in the absence of labeled data. Handling \textit{domain-shift} and \textit{category-shift} simultaneously
is necessary for any \textit{open-set} DA algorithm, which is not supported by these methods.

\vspace{2mm}

\noindent Our formulation of an \textit{inheritable} model for \textit{open-set} DA is much different from prior arts - not only is it robust to \textit{negative-transfer} but also facilitates domain adaptation in the absence of data-exchange.

\section{Unsupervised Open-Set Domain Adaptation}
\label{sec:osda}

In this section, we formally define the \textit{vendor-client} paradigm and \textit{inheritability} in the context of unsupervised \textit{open-set} domain adaptation (UODA).

\subsection{Preliminaries}
\vspace{-1mm}
\noindent \textbf{Notation.} Given an input space $\mathcal{X}$ and output space $\mathcal{Y}$, the {source} and target domains are characterized by the distributions $p$ and $q$ on $\mathcal{X} \times \mathcal{Y}$ respectively. Let $p_{x}$, $q_{x}$ denote the marginal input distributions and $p_{y|x}, q_{y|x}$ denote the conditional output distribution of the two domains. Let $\mathcal{C}_s, \mathcal{C}_t \subset \mathcal{Y}$ denote the respective label sets for the classification tasks ($\mathcal{C}_s \subset \mathcal{C}_t$). In the UODA problem, a labeled {source} dataset $\mathcal{D}_s = \{(x_s, y_s):x_s\sim{p_{x}}, y_s\sim{p_{y|x}}\}$ and an unlabeled {target} dataset $\mathcal{D}_t = \{x_t:x_t\sim{q_{x}}\}$ are considered. The goal is to assign a label for each {target} instance $x_t$, by predicting the class for those in shared classes ($\mathcal{C}_t^{sh} = \mathcal{C}_s$), and an `\textit{unknown}' label for those in unshared classes ($\mathcal{C}_t^{uk} = \mathcal{C}_t \setminus \mathcal{C}_s$). For simplicity, we denote the distributions of target-shared and target-\textit{unknown} instances as $q^{sh}$ and $q^{uk}$ respectively. We denote the model trained on the {source} domain as $h_s$ ({source} predictor) and the model adapted to the {target} domain as $h_t$ ({target} predictor).


\vspace{1mm}

\noindent \textbf{Performance Measure.} The primary goal of UODA is to improve the performance on the {target} domain. Hence, the performance of any UODA algorithm is measured by the error rate of target predictor $h_t$, \ie $\xi_{q}(h_t)$ which is empirically estimated as $\hat{\xi}_{q}(h_t) = \mathbb{P}_{\{(x_t, y_t) \sim q\}} [h_t(x_t) \neq y_t]$, where $\mathbb{P}$ is the probability estimated over the instances $\mathcal{D}_t$.  





\subsection{The vendor-client paradigm}


\vspace{-1mm}

The central focus of our work is to realize a practical DA paradigm which is fundamentally viable in the absence of the co-existance of the source and target domains. With this intent, we formalize our DA paradigm.



\vspace{1mm}

\noindent \textbf{Definition 1} (\textit{vendor-client} paradigm). \textit{Consider a \textit{vendor} with access to a labeled {source} dataset $\mathcal{D}_s$ 
and a \textit{client} having unlabeled instances $\mathcal{D}_t$ sampled from the {target} domain. In the \textit{vendor-client} paradigm, the \textit{vendor} learns a {source} predictor $h_s$ using $\mathcal{D}_s$ to model the conditional $p_{y|x}$, and shares $h_s$ to the \textit{client}.
Using $h_s$ and $\mathcal{D}_t$, the \textit{client} learns a {target} predictor $h_t$ to model the conditional $q_{y|x}$.}

\vspace{1mm}

This paradigm satisfies the two important properties; 1) it does not assume data-exchange between the \textit{vendor} and the \textit{client} which is fundamental to cope up with the dynamically reforming digital privacy and copyright regulations 
and, 2) a single \textit{vendor} model can be shared with multiple \textit{clients} thereby minimizing the effort spent on {source} training. Thus, this paradigm has a greater practical significance than the traditional UDA setup where each adaptation step requires an additional supervision from the source data~\cite{sta_open_set,saito2018open}.
Following this paradigm, our goal is to realize the conditions on which one can successfully learn a {target} predictor. To this end, we formalize the \textit{inheritability} of 
\textit{task-specific} knowledge of the 
source-trained model.


\subsection{Inheritability}

\vspace{-1mm}

We define an \textit{inheritable} model from the perspective of learning a predictor ($h_t$) for the target task. Intuitively, given a hypothesis class $\mathcal{H} \subseteq \{h ~|~h:\mathcal{X}\rightarrow\mathcal{Y}\}$, an \textit{inheritable} model $h_s$ should be sufficient (\ie in the absence of source domain data) to learn a {target} predictor $h_t$ whose performance is close to that of the best predictor in $\mathcal{H}$. 

\vspace{1mm}

\noindent \textbf{Definition 2} (Inheritability criterion). 
\textit{Let $\mathcal{H} \subseteq \{h ~|~h:\mathcal{X}\rightarrow\mathcal{Y}\}$ be a hypothesis class, $\epsilon > 0$, and $\delta \in (0,1)$. A {source} predictor $h_s : \mathcal{X} \rightarrow \mathcal{Y}$ is termed \textit{inheritable} relative to the hypothesis class $\mathcal{H}$, if a {target} predictor $h_t : \mathcal{X} \rightarrow \mathcal{Y}$ can be learned using an unlabeled {target} sample $\mathcal{D}_t=\{x_t~:~x_t\sim{q_x}\}$ when given access to the parameters of $h_s$, such that, with probability at least $(1-\delta)$ the target error of $h_t$ does not exceed that of the best predictor in $\mathcal{H}$ by more than $\epsilon$. Formally,}

\vspace{-2mm}

\begin{equation}
    \label{eq:inheritability_bound}
    \mathbb{P}[\xi_{q}(h_t) \leq \xi_{q}(\mathcal{H}) + \epsilon~|~h_s, \mathcal{D}_t] \geq 1 - \delta
\end{equation}{}
\vspace{-4mm}

\noindent where, $\xi_{q}(\mathcal{H}) = \min_{h\in\mathcal{H}} \xi_{q}(h)$ and $\mathbb{P}$ is computed over the choice of sample $\mathcal{D}_t$. This definition suggests that an \textit{inheritable} model 
is capable of reliably transferring the \textit{task-specific} knowledge to the target domain in the absence of the source data, which is necessary for the \textit{vendor-client} paradigm. 
Given this definition, a natural question is,
how to quantify \textit{inheritability} of a \textit{vendor} model for the {target} task. In the next Section, we address this question by demonstrating the design of \textit{inheritable} models for UODA.


 



\section{Approach}
\label{sec:Approach}

How to design \textit{inheritable} models? There can be several ways, depending upon the \textit{task-specific} knowledge required by the \textit{client}. For instance, in UODA, the \textit{client} must effectively learn a classifier in the presence of both \textit{domain-shift} and \textit{category-shift}.
Here, not only is the knowledge of class-separability essential, but also the ability to detect new target categories as \textit{unknown} is vital to avoid \textit{negative-transfer}.
By effectively identifying such challenges, one can develop \textit{inheritable} models for tasks that require
\textit{vendor's} dataset. Here, we demonstrate UODA using an \textit{inheritable} model.

\begin{figure}
    \centering
    \includegraphics[width=0.84\linewidth]{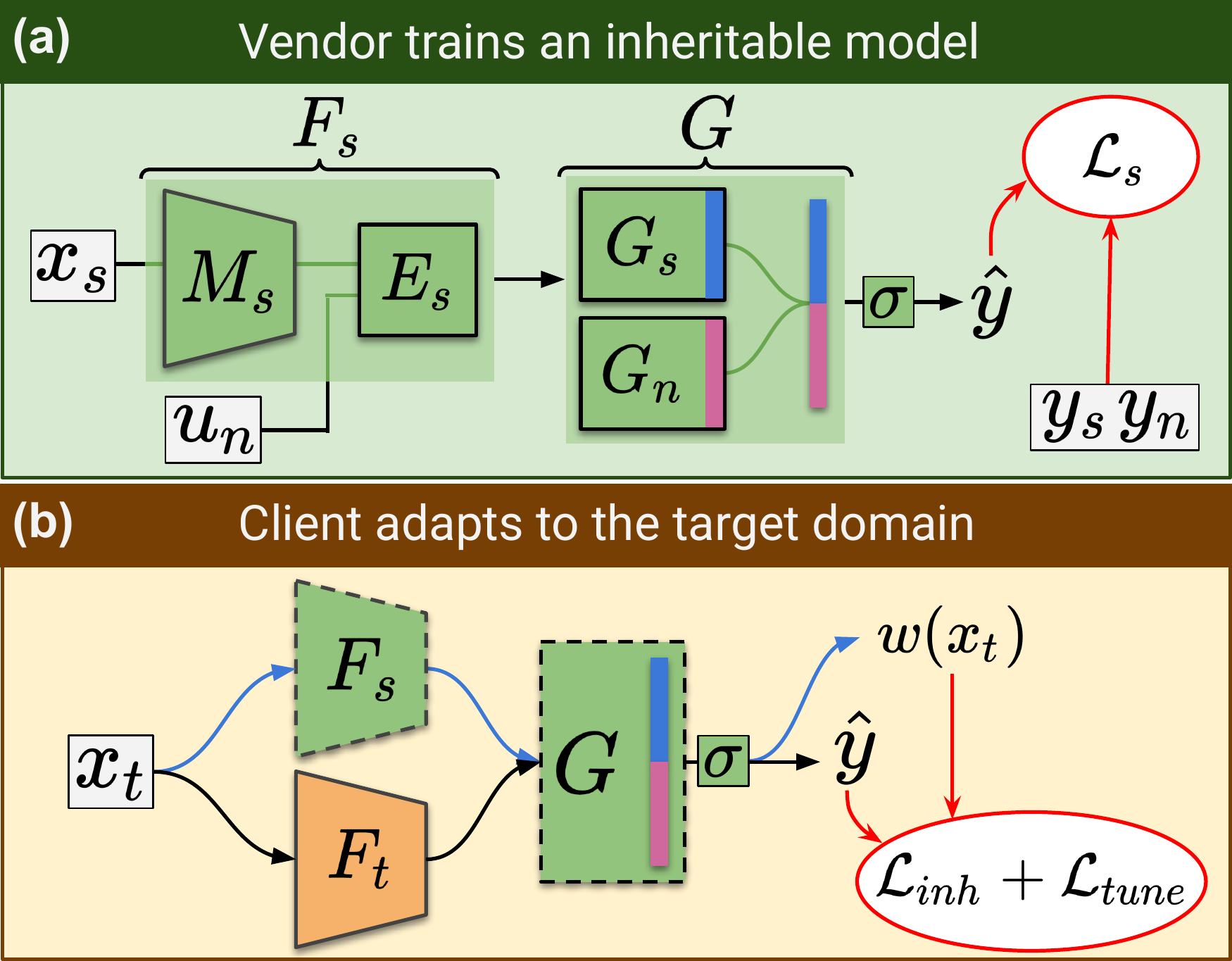}
    \caption{\small The architectures for \textbf{A)} \textit{vendor}-side training and \textbf{B)} \textit{client}-side adaptation. Dashed border denotes a frozen network.}
    \label{fig:architecture}
    \vspace{-4mm}
\end{figure}


\begin{figure*}
    \centering
    \includegraphics[width=0.95\linewidth]{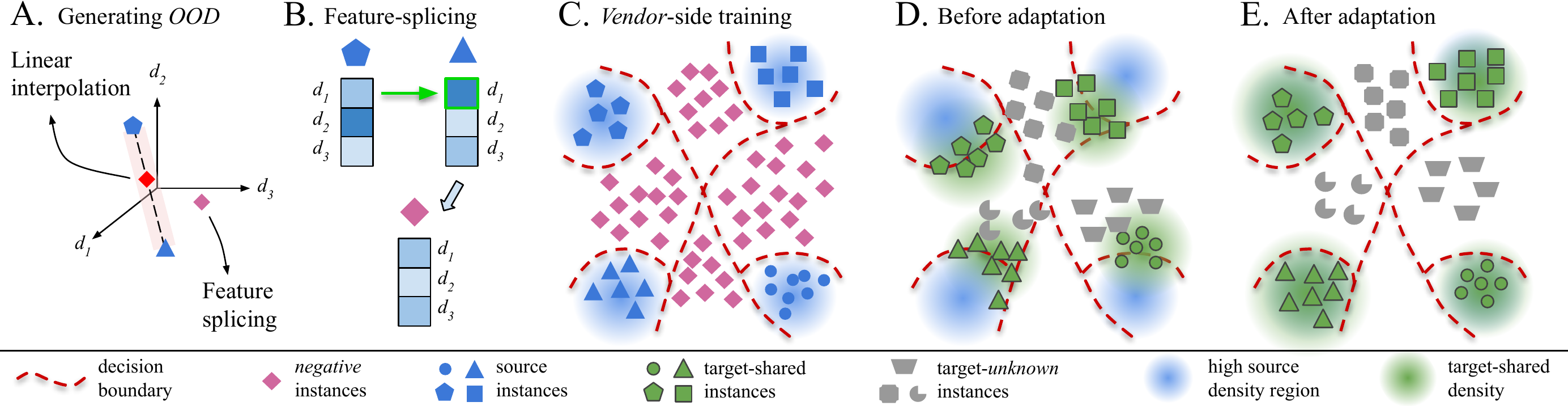}
    \caption{\textbf{A)} An example of a \textit{negative} instance generated in a 3-dimensional space ($d_1,d_2,d_3$) using linear interpolation and Feature-splicing. \textbf{B)} Feature-splicing by suppressing the class-discriminative traits (here, we replace the top-(1/3) percentile activation, $d_1$). \textbf{C)} An \textit{inheritable} model with negative classes. \textbf{D)} \textit{Domain-shift} before adaptation. \textbf{E)} Successful adaptation. Best viewed in color.}
    \label{fig:negative_clustering}
    \vspace{-3mm}
\end{figure*}

\subsection{Vendor trains an inheritable model}
\label{sec:vendor_trains}

\vspace{-1mm}

In UODA, the primary challenge is to tackle \textit{negative-transfer}. This challenge arises due to the overconfidence issue \cite{lee2018training_ood} in deep models, where \textit{unknown} target instances are confidently predicted into the shared classes, and thus get aligned with the source domain. 
Methods such as \cite{UDA_2019_CVPR} tend to avoid \textit{negative-transfer} by leveraging a domain discriminator to assign a low instance-level weight for potentially \textit{unknown} target instances during adaptation.
However, solutions such as a domain discriminator are infeasible in the absence of data-exchange between the \textit{vendor} and the \textit{client}. 
Thus, an \textit{inheritable} model should have the ability to characterize the {source} distribution, which will facilitate the detection of \textit{unknown} target instances during adaptation. Following this intuition, we design the architecture.

\vspace{1mm}

\noindent \textbf{a) Architecture.} As shown in Fig.~\ref{fig:architecture}{\color{red}A}, the feature extractor $F_s$ comprises of a backbone CNN model $M_s$ and fully connected layers $E_s$. The classifier $G$ contains two sub-modules, a source classifier $G_s$ with $|\mathcal{C}_s|$ classes, and an auxiliary \textit{out-of-distribution} (OOD) classifier $G_n$ with $K$ classes accounting for the `\textit{negative}' region not covered by the source distribution (Fig. \ref{fig:negative_clustering}{\color{red}C}). The output $\hat{y}_s$ for each input $x_s$ is obtained by concatenating the outputs of $G_s$ and $G_n$ (\ie concatenating $G_s(F_s(x_s))$ and $G_n(F_s(x_s))$) followed by softmax activation. 
This equips the model with the ability to capture the class-separability knowledge (in $G_s$) and to detect OOD instances (via $G_n$). 
This setup is motivated by the fact that the overconfidence issue can be addressed by minimizing the classifier's confidence for OOD instances \cite{lee2018training_ood}.
Accordingly, the confidence of $G_s$ is maximized for \textit{in-distribution} (source) instances, and minimized for OOD instances (by maximizing the confidence of $G_n$). 

\vspace{1mm}

\noindent \textbf{b) Dataset preparation.}
To effectively learn OOD detection, we augment the source dataset with synthetically generated \textit{negative} instances, \ie $\mathcal{D}_n = \{(u_n, y_n) : u_n\sim{r_u}, y_n\sim{r_{y|u}}\}$, where $r_u$ and $r_{y|u}$ are the marginal latent space distribution and the conditional output distribution of the \textit{negative} instances respectively.
We use $\mathcal{D}_n$, to model the low source-density region as \textit{out-of-distribution} (see Fig.~\ref{fig:negative_clustering}{\color{red}C}). 
To obtain $\mathcal{D}_n$, a possible approach explored by \cite{lee2018training_ood} could be to use a GAN framework to generate `boundary' samples. However, this is computationally intensive and introduces additional parameters for training. Further, we require these \textit{negative} samples to cover a large portion of the OOD region. This eliminates a direct use of linear interpolation techniques such as \textit{mixup}~\cite{mixup,verma2019manifoldmixup} which result in features generated within a restricted region (see Fig. \ref{fig:negative_clustering}{\color{red}A}). Indeed, we propose an efficient way to generate OOD samples, which we call as the feature-splicing technique.

\textbf{Feature-splicing.} 
It is widely known that in deep CNNs, higher convolutional layers specialize in capturing class-discriminative properties \cite{zeiler2014visualizingandunderstandingcnn}.
For instance, \cite{zhang2018interpretable} assigns each filter in a high conv-layer with an object part, demonstrating that each filter learns a different \textit{class-specific} trait.
As a result of this specificity, especially when a rectified activation function (\eg ReLU) is used, feature maps receive a high activation whenever the learned \textit{class-specific} trait is observed in the input \cite{Chen_2019_ICCV_DAFL}.
Consequently, we argue that, by suppressing such high activations, we obtain features devoid of the properties specific to the source classes and hence would more accurately represent the OOD samples. 
Then, enforcing a low classifier confidence for these samples can mitigate the overconfidence issue.

Feature-splicing is performed by replacing the top-$d$ percentile activations, at a particular feature layer, with the corresponding activations pertaining to an instance belonging to a different class (see Fig.~\ref{fig:negative_clustering}{\color{red}B}).  Formally,

\vspace{-3.5mm}
\begin{equation}
    u_n = \phi_{d}(u_s^{c_i}, u_s^{c_j}) ~~\text{for}~~ c_i,c_j \in \mathcal{C}_s, {c_i}\neq{c_j}
\end{equation}
\vspace{-4.5mm}

where, $u_s^{c_i} = M_s(x_s^{c_i})$ for a source image $x_s^{c_i}$ belonging to class $c_i$, and $\phi_d$ is the feature-splicing operator which replaces the top-$d$ percentile activations in the feature $u_s^{c_i}$ with the corresponding activations in $u_s^{c_j}$ as shown in Fig. \ref{fig:negative_clustering}{\color{red}B} (see Suppl. for algorithm). 
This process results in a feature which is devoid of the \textit{class-specific} traits, but lies near the source distribution.
To label these \textit{negative} instances, we perform a $K$-means clustering and assign a unique \textit{negative} class label to each cluster of samples. By training the auxiliary classifier $G_n$ to discriminate these samples into these $K$ \textit{negative} classes, we mitigate the overconfidence issue as stated earlier. We found feature-splicing to be effective in practice. See Suppl. for other techniques that we explored.

\vspace{1mm}

\noindent \textbf{c) Training procedure.} We train the model in two steps. First, we pre-train $\{F_s, G_s\}$ using source data $\mathcal{D}_s$ by employing the standard cross-entropy loss,

\vspace{-2.5mm}
\begin{equation}
    \mathcal{L}_{b} = \mathcal{L}_{CE}(\sigma(G_s(F_s(x_s))), y_s)
\end{equation}
\vspace{-4.5mm}

where, $\sigma$ is the softmax activation function. Next, we freeze the backbone model $M_s$, and generate \textit{negative} instances $\mathcal{D}_n=\{(u_n, y_n)\}$ by performing feature-splicing using source features at the last layer of $M_s$. We then continue the training of the modules $\{E_s, G_s, G_n\}$ using supervision from both $\mathcal{D}_s$ and $\mathcal{D}_n$,

\vspace{-2.5mm}
\begin{equation}
    \label{eq:vendor_loss}
    \mathcal{L}_{s} = \mathcal{L}_{CE}(\hat{y}_s, y_s) + \mathcal{L}_{CE}(\hat{y}_n, y_n)
\end{equation}

where, $\hat{y}_s = \sigma(G(F_s(x_s)))$ and $\hat{y}_n = \sigma(G(E_s(u_n)))$, and the output of $G$ is obtained as described in Sec.~\ref{sec:vendor_trains}{\color{red}a} (and depicted in Fig.~\ref{fig:architecture}). The joint training of $G_s$ and $G_n$, allows the model to capture the class-separability knowledge (in $G_s$) while characterizing the \textit{negative} region (in $G_n$), which renders a superior knowledge \textit{inheritability}.
Once the \textit{inheritable} model $h_s=\{F_s,G\}$ is trained, it is shared to the \textit{client} for performing UODA.

\subsection{Client adapts to the target domain} 
\label{sec:client_adapts}

\vspace{-1mm}


With a trained \textit{inheritable} model ($h_s$) in hand, the first task is to measure the degree of \textit{domain-shift} to determine the \textit{inheritability} of the \textit{vendor's} model. This is followed by a selective adaptation procedure which encourages shared classes to align while avoiding \textit{negative-transfer}.


\vspace{1mm}

\noindent \textbf{a) Quantifying inheritability.} 
In presence of a small \textit{domain-shift}, 
most of the 
target-shared instances (pertaining to classes in $\mathcal{C}_t^{sh}$)
will lie close to the high source-density regions in the latent space (\eg Fig.~\ref{fig:negative_clustering}{\color{red}E}). Thus, one can rely on the class-separability knowledge of $h_s$ to predict target labels. However, this knowledge becomes less reliable with increasing \textit{domain-shift} as the concentration of target-shared instances near the high density regions decreases (\eg Fig.~\ref{fig:negative_clustering}{\color{red}D}). Thus, the \textit{inheritability} of $h_s$ for the target task would decrease with increasing \textit{domain-shift}. Moreover, target-\textit{unknown} instances (pertaining to classes in $\mathcal{C}_t^{uk}$) are more likely to lie in the low source-density region than target-shared instances. With this intuition, we define an \textit{inheritability} metric $w$ which satisfies,

\vspace{-5mm}

\begin{equation}
    \label{eq:inheritability_inequality}
    \expectation_{x_s \sim p_x} w(x_s) \ge \expectation_{x_t \sim q_x^{sh}} w(x_t) \ge \expectation_{x_t \sim q_x^{uk} } w(x_t)
\end{equation}

\vspace{-1mm}

We leverage the classifier confidence to realize an instance-level measure of \textit{inheritability} as follows,

\vspace{-3mm}

\begin{equation}
    \label{eq:instance_level_inheritability}
    w(x) = \max_{c_i \in \mathcal{C}_s} ~[\sigma(G(F_s(x)))]_{c_i}
\end{equation}

\vspace{-1mm}

where $\sigma$ is the softmax activation function. Note that although softmax is applied over the entire output of $G$, $\max$ is evaluated over those corresponding to $G_s$ (shaded in blue in Fig.~\ref{fig:architecture}). We hypothesize that this measure follows Eq.~\ref{eq:inheritability_inequality}, since, the source instances (in the high density region) receive the highest $G_s$ confidence, followed by target-shared instances (some of which are away from the high density region), while the target-\textit{unknown} instances receive the least confidence (many of which lie away from the high density regions). Extending the instance-level \textit{inheritability}, we define a model \textit{inheritability} over the entire target dataset as,

\vspace{-2mm}

\begin{equation}
    \label{eq:model_inheritability}
    \mathcal{I}(h_s, \mathcal{D}_s, \mathcal{D}_t) = \dfrac{\operatorname{mean}_{x_t \in \mathcal{D}_t}{w(x_t)}}{\operatorname{mean}_{x_s \in \mathcal{D}_s}{w(x_s)}}
\end{equation}


A higher $\mathcal{I}$ arises from a smaller \textit{domain-shift} implying a greater \textit{inheritability} of \textit{task-specific} knowledge (\eg class-separability for UODA) to the target domain. Note that $\mathcal{I}$ is a constant for a given triplet $\{h_s, \mathcal{D}_s, \mathcal{D}_t\}$ and the value of the denominator in Eq.~\ref{eq:model_inheritability} can be obtained from the \textit{vendor}.




\vspace{1mm}

\noindent \textbf{b) Adaptation procedure.} For performing adaptation to the target domain, we learn a target-specific feature extractor $F_t=\{M_t, E_t\}$ as shown in Fig.~\ref{fig:architecture}{\color{red}B} (similar in architecture to $F_s$). $F_t$ is initialized from the source feature extractor $F_s = \{M_s, E_s\}$, and is gradually trained to selectively align the shared classes in the pre-classifier space (input to $G$) to avoid \textit{negative-transfer}. The adaptation involves two processes - \textit{inherit} (to acquire the class-separability knowledge) and \textit{tune} (to avoid \textit{negative-transfer}).


\textbf{Inherit.} As described in Sec.~\ref{sec:client_adapts}{\color{red}a}, the class-separability knowledge of $h_s$ is reliable for target samples with high $w$. Subsequently, we choose top-$k$ percentile target instances based on $w(x_t)$ and obtain pseudo-labels using the source model, $y_p = \operatorname{argmax}_{c_i \in \mathcal{C}_s}~[\sigma(G(F_s(x_t)))]_{c_i}$. 
Using the cross-entropy loss we enforce the target predictions to match the pseudo-labels for these instances, thereby \textit{inheriting} the class-separability knowledge,

\vspace{-2mm}

\begin{equation}
    \label{loss_inherit}
    \mathcal{L}_{inh} = \mathcal{L}_{CE}(\sigma(G(F_t(x_t))), y_p)
\end{equation}{}

\vspace{-4mm}

\textbf{Tune.} 
In the absence of label information, entropy minimization \cite{long2016unsupervised,grandvalet2005semi} is popularly employed to move the features of unlabeled instances towards the high confidence regions. However, to avoid \textit{negative-transfer}, 
instead of a direct application of entropy minimization,
we use $w$ as a soft instance weight in our loss formulation. Target instances with higher $w$ are guided towards the high source density regions, while those with lower $w$ are pushed into the \textit{negative} regions (see Fig.~\ref{fig:negative_clustering}{\color{red}{D}$\rightarrow${E}}). This separation is a key to minimize the effect of \textit{negative-transfer}.

On a coarse level, using the classifier $G$ we obtain the probability $\hat{s}$ that an instance belongs to the shared classes as $\hat{s} = \sum_{c_i \in \mathcal{C}_s} [\sigma(G(F_t(x_t)))]_{c_i}$. Optimizing the following loss encourages a separation of shared and \textit{unknown} classes,

\vspace{-5mm}

\begin{equation}
    \mathcal{L}_{t1} = - w(x_t) \operatorname{log}(\hat{s}) - (1-w(x_t)) \operatorname{log}(1-\hat{s}) 
\end{equation}

\vspace{-1mm}

To further encourage the alignment of shared classes on a fine level, we separately calculate probability vectors for $G_s$ as, $z_t^{sh} = \sigma(G_s(F_t(x_t)))$, and for $G_n$ as, $z_t^{uk} = \sigma(G_n(F_t(x_t)))$, and minimize the following loss, 


\vspace{-5mm}

\begin{equation}
    \mathcal{L}_{t2} = w(x_t) \operatorname{H}(z_t^{sh}) + (1-w(x_t)) \operatorname{H}(z_t^{uk})
\end{equation}




\noindent where, $\operatorname{H}$ is the Shannon's entropy. The total loss $\mathcal{L}_{tune} = \mathcal{L}_{t1} + \mathcal{L}_{t2}$ selectively aligns the shared classes, while avoiding \textit{negative-transfer}. Thus, the final adaptation loss is,

\vspace{-3mm}

\begin{equation}
    \mathcal{L}_a = \mathcal{L}_{inh} + \mathcal{L}_{tune}
\end{equation}






We now present a discussion on the success of this adaptation procedure from the theoretical perspective.

\subsection{Theoretical Insights}
\label{theoretical_analysis}

\vspace{-1mm}

We defined the \textit{inheritability} criterion in Eq.~\ref{eq:inheritability_bound} for transferring the \textit{task-specific} knowledge to the target domain. To show that the knowledge of class-separability is indeed \textit{inheritable}, it is sufficient to demonstrate that the \textit{inheritability} criterion holds for the shared classes. Extending Theorem {\color{red}3} in \cite{ben2010theory}, we obtain the following result.

\vspace{1mm}


\noindent \textbf{Result 1.} \textit{Let $\mathcal{H}$ be a hypothesis class of VC dimension $d$. Let $\mathcal{S}$ be a labeled sample set of $m$ points drawn from $q^{sh}$. If 
$\widehat{h}_t \in \mathcal{H}$
be the empirical minimizer of $\xi_{q^{sh}}$ on $\mathcal{S}$, and $h_t^{*} = \operatorname{argmin}_{h\in\mathcal{H}}\xi_{q^{sh}}(h)$ be the optimal hypothesis for $q^{sh}$, then for any $\delta \in (0, 1)$, we have with probability of at least $1-\delta$ (over the choice of samples),}

\vspace{-5mm}

\begin{equation}
    \label{result_1}
    \xi_{q^{sh}}(\widehat{h}_t) \leq \xi_{q^{sh}}(h^{*}_t) + 4\sqrt{\dfrac{2d\operatorname{log}(2(m+1)) + 2\operatorname{log}(8/\delta)}{m}}
\end{equation}



See Supplementary for the derivation of this result.
Essentially, using $m$ labeled target-shared instances, one can train a predictor (here, $\widehat{h}_t$) which satisfies Eq.~\ref{result_1}.
However, in a completely unsupervised setting, the only way to obtain target labels is to exploit the knowledge of the \textit{vendor's} model.
This is precisely what the pseudo-labeling process achieves. Using an \textit{inheritable} model ($h_s$), we pseudo-label the
top-$k$ percentile target instances with high precision and enforce $\mathcal{L}_{inh}$.
In doing so, we condition the target model to satisfy Eq.~\ref{result_1}, which is the \textit{inheritability} criterion for shared categories (given unlabeled instances $\mathcal{D}_t$ and source model $h_s$). 
Thus, the knowledge of class-separability is transferred to the target model during the adaptation process.

Note that, with increasing number of labeled target instances (increasing $m$), the last term in Eq.~\ref{result_1} decreases. In our formulation, this is achieved by enforcing $\mathcal{L}_{tune}$, which can be regarded as a way to self-supervise the target model. In Sec.~\ref{sec:experiments} we verify that, during adaptation the precision of target predictions improves over time. This self-supervision with an increasing number of correct labels is, in effect, similar to having a larger sample size $m$ in Eq.~\ref{result_1}. Thus, adaptation tightens the bound in Eq.~\ref{result_1} (see Suppl.).

\vspace{-1mm}
\section{Experiments}
\label{sec:experiments}

\vspace{-1mm}
In this section, we evaluate the performance of unsupervised \textit{open-set} domain adaptation using \textit{inheritable} models.

\begin{table*}[t]
    \setlength{\tabcolsep}{5pt}
    \centering
    \caption{\small Results on \textbf{Office-31} (ResNet-50). $|\mathcal{C}_s|=10$, $|\mathcal{C}_t|=20$. \textit{Ours} denotes adaptation using an \textit{inheritable} model following the \textit{vendor-client} paradigm, while all other methods use source domain data during adaptation.
    }
    \resizebox{1\textwidth}{!}{%
        
        \begin{tabular}{lcccccccccccccccc}
            \hline
            \hline
            \multirow{2}{*}{Method} & \multicolumn{2}{c}{A$\rightarrow$W} & \multicolumn{2}{c}{A$\rightarrow$D} & \multicolumn{2}{c}{D$\rightarrow$W} & \multicolumn{2}{c}{W$\rightarrow$D} & \multicolumn{2}{c}{D$\rightarrow$A} & \multicolumn{2}{c}{W$\rightarrow$A} & \multicolumn{2}{c}{Avg
            } \\
            \cmidrule(lr){2-15}
            & OS & OS* & OS & OS* & OS & OS* & OS & OS* & OS & OS* & OS & OS* & OS & OS* \\
            \hline
            ResNet & 82.5$\pm$1.2 & 82.7$\pm$0.9 & 85.2$\pm$0.3 & 85.5$\pm$0.9 & 94.1$\pm$0.3 & 94.3$\pm$0.7 & 96.6$\pm$0.2 & 97.0$\pm$0.4 & 71.6$\pm$1.0 & 71.5$\pm$1.1 & 75.5$\pm$1.0 & 75.2$\pm$1.6 & 84.2 & 84.4 \\
            RTN \cite{long2016unsupervised} & 85.6$\pm$1.2 & 88.1$\pm$1.0 & 89.5$\pm$1.4 & 90.1$\pm$1.6 & 94.8$\pm$0.3 & 96.2$\pm$0.7 & 97.1$\pm$0.2 & 98.7$\pm$0.9 & 72.3$\pm$0.9 & 72.8$\pm$1.5 & 73.5$\pm$0.6 & 73.9$\pm$1.4 & 85.4 & 86.8 \\
            DANN \cite{ganin2016domain} & 85.3$\pm$0.7 & 87.7$\pm$1.1 & 86.5$\pm$0.6 & 87.7$\pm$0.6 & 97.5$\pm$0.2 & \textbf{98.3}$\pm$0.5 & 99.5$\pm$0.1 & \textbf{100.0}$\pm$.0 & 75.7$\pm$1.6 & 76.2$\pm$0.9 & 74.9$\pm$1.2 & 75.6$\pm$0.8 & 86.6 & 87.6 \\
            OpenMax \cite{openSetbendale2016towards} & 87.4$\pm$0.5 & 87.5$\pm$0.3 & 87.1$\pm$0.9 & 88.4$\pm$0.9 & 96.1$\pm$0.4 & 96.2$\pm$0.3 & 98.4$\pm$0.3 & 98.5$\pm$0.3 & 83.4$\pm$1.0 & 82.1$\pm$0.6 & 82.8$\pm$0.9 & 82.8$\pm$0.6 & 89.0 & 89.3 \\
            ATI-$\lambda$ \cite{panareda2017open} &  87.4$\pm$1.5 & 88.9$\pm$1.4 & 84.3$\pm$1.2 & 86.6$\pm$1.1 & 93.6$\pm$1.0 & 95.3$\pm$1.0 & 96.5$\pm$0.9 & 98.7$\pm$0.8 & 78.0$\pm$1.8 & 79.6$\pm$1.5 & 80.4$\pm$1.4 & 81.4$\pm$1.2 & 86.7 & 88.4 \\
            OSBP \cite{saito2018open} & 86.5$\pm$2.0 & 87.6$\pm$2.1 & 88.6$\pm$1.4 & 89.2$\pm$1.3 & 97.0$\pm$1.0 & 96.5$\pm$0.4 & 97.9$\pm$0.9 & 98.7$\pm$0.6 & 88.9$\pm$2.5 & 90.6$\pm$2.3 & 85.8$\pm$2.5 & 84.9$\pm$1.3 & 90.8 & 91.3 \\
            STA \cite{sta_open_set} & 89.5$\pm$0.6 & 92.1$\pm$0.5 & 93.7$\pm$1.5 & 96.1$\pm$ 0.4 & \textbf{97.5}$\pm$0.2 & 96.5$\pm$0.5 & 99.5$\pm$0.2 & 99.6$\pm$0.1 & 89.1$\pm$0.5 & \textbf{93.5}$\pm$0.8 & 87.9$\pm$0.9 & 87.4$\pm$0.6 & 92.9 & 94.1 \\
            \hline
            \textit{Ours} & \textbf{91.3}$\pm$0.7 & \textbf{93.2}$\pm$1.2 & \textbf{94.2}$\pm$1.1 & \textbf{97.1}$\pm$0.8 & 96.5$\pm$0.5 & 97.4$\pm$0.7 & \textbf{99.5}$\pm$0.2 & 99.4$\pm$0.3 & \textbf{90.1}$\pm$0.2 & 91.5$\pm$ 0.2 & \textbf{88.7}$\pm$1.3 & \textbf{88.1}$\pm$0.9 & \textbf{93.4} & \textbf{94.5} \\
            \hline
            \hline
        \end{tabular}
    }
    
    \label{tab:exp_office31}
\end{table*}

\begin{table*}[t]
    \vspace{-2mm}
    \centering
    \caption{Results on \textbf{Office-Home} (ResNet-50). $|\mathcal{C}_s|=25$, $|\mathcal{C}_t|=65$. \textit{Ours} denotes adaptation using an \textit{inheritable} model.
    }
    \resizebox{1\textwidth}{!}{%
        \begin{tabular}{lccccccccccccccc}
            \hline
            \hline
            Method & Ar$\rightarrow$Cl & Pr$\rightarrow$Cl & Rw$\rightarrow$Cl & Ar$\rightarrow$Pr & Cl$\rightarrow$Pr & Rw$\rightarrow$Pr & Cl$\rightarrow$Ar & Pr$\rightarrow$Ar & Rw$\rightarrow$Ar & Ar$\rightarrow$Rw & Cl$\rightarrow$Rw & Pr$\rightarrow$Rw & Avg \\
            \hline
            ResNet & 53.4$\pm$0.4 & 52.7$\pm$0.6 & 51.9$\pm$0.5 & 69.3$\pm$0.7 & 61.8$\pm$0.5 & 74.1$\pm$0.4 & 61.4$\pm$0.6 & 64.0$\pm$0.3 & 70.0$\pm$0.3 & 78.7$\pm$0.6 & 71.0$\pm$0.6 & 74.9$\pm$0.9 & 65.3 \\
            ATI-$\lambda$ \cite{panareda2017open} & 55.2$\pm$1.2 & 52.6$\pm$1.6 & 53.5$\pm$1.4 & 69.1$\pm$1.1 & 63.5$\pm$1.5 & 74.1$\pm$1.5 & 61.7$\pm$1.2 & 64.5$\pm$0.9 & 70.7$\pm$0.5 & 79.2$\pm$0.7 & 72.9$\pm$0.7 & 75.8$\pm$1.6 & 66.1 \\
            DANN \cite{ganin2016domain}& 54.6$\pm$0.7 & 49.7$\pm$1.6 & 51.9$\pm$1.4 & 69.5$\pm$1.1 & 63.5$\pm$1.0 & 72.9$\pm$0.8 & 61.9$\pm$1.2 & 63.3$\pm$1.0 & 71.3$\pm$1.0 & 80.2$\pm$0.8 & 71.7$\pm$0.4 & 74.2$\pm$0.4 & 65.4 \\
            OSBP \cite{saito2018open} & 56.7$\pm$1.9 & 51.5$\pm$2.1 & 49.2$\pm$2.4 & 67.5$\pm$1.5 & 65.5$\pm$1.5 & 74.0$\pm$1.5 & 62.5$\pm$2.0 & 64.8$\pm$1.1 & 69.3$\pm$1.1 & 80.6$\pm$0.9 & 74.7$\pm$2.2 & 71.5$\pm$1.9 & 65.7 \\
            OpenMax \cite{openSetbendale2016towards}& 56.5$\pm$0.4 & 52.9$\pm$0.7 & 53.7$\pm$0.4 & 69.1$\pm$0.3 & 64.8$\pm$0.4 & 74.5$\pm$0.6 & \textbf{64.1}$\pm$0.9 & 64.0$\pm$0.8 & 71.2$\pm$0.8 & 80.3$\pm$0.8 & 73.0$\pm$0.5 & 76.9$\pm$0.3 & 66.7 \\
            STA \cite{sta_open_set} &  58.1$\pm$0.6 & 53.1$\pm$0.9 & 54.4$\pm$1.0 & \textbf{71.6}$\pm$1.2 & 69.3$\pm$1.0 & \textbf{81.9}$\pm$0.5 & 63.4$\pm$0.5 & 65.2$\pm$0.8 & 74.9$\pm$1.0 & \textbf{85.0}$\pm$0.2 & \textbf{75.8}$\pm$0.4 & 80.8$\pm$0.3 & 69.5 \\
            \hline
            \textit{Ours} & \textbf{60.1}$\pm$0.7 & \textbf{54.2}$\pm$1.0&\textbf{56.2}$\pm$1.7&70.9$\pm$1.4 & \textbf{70.0}$\pm$1.7&78.6$\pm$0.6&64.0$\pm$0.6&\textbf{66.1}$\pm$1.3&\textbf{74.9}$\pm$0.9 & 83.2$\pm$0.9 & 75.7$\pm$1.3&\textbf{81.3}$\pm$1.4 & \textbf{69.6} \\
            \hline
            \hline
        \end{tabular}
    }
    
    \label{tab:exp_officehome}
    \vspace{-1mm}
\end{table*}

\begin{table}[t]
    \vspace{-3mm}
    \centering
    \caption{Results on \textbf{VisDA} (VGGNet). $|\mathcal{C}_s|=6$, $|\mathcal{C}_t|=12$. \textit{Ours} denotes adaptation using an \textit{inheritable} model.
    }
    \setlength{\tabcolsep}{5pt}
    \resizebox{1\columnwidth}{!}{%
    \begin{tabular}{lccccccccc}
        \hline
        \hline
         \multirow{2}{*}{Method} & \multicolumn{8}{c}{Synthetic $\rightarrow$ Real} \\
         \cmidrule(lr){2-9}
         & bicycle & bus & car & m-cycle & train & truck & OS & OS* \\
         \hline
         OSVM \cite{openSetjain2014multi} & 31.7 & 51.6 & 66.5 & 70.4 & \textbf{88.5} & 20.8 & 52.5 & 54.9 \\
        MMD+OSVM & 39.0 & 50.1 & 64.2 & 79.9 & 86.6 & 16.3 & 54.4 & 56.0 \\
        DANN+OSVM & 31.8 & 56.6 & \textbf{71.7} & 77.4 & 87.0 & 22.3 & 55.5 & 57.8 \\
        ATI-$\lambda$ \cite{panareda2017open} & 46.2 & 57.5 & 56.9 & 79.1 & 81.6 & \textbf{32.7} & 59.9 & 59.0 \\
        OSBP \cite{saito2018open} &  51.1 & 67.1 & 42.8 & 84.2 & 81.8 & 28.0 &  62.9 & 59.2 \\
        STA \cite{sta_open_set} & 52.4 & \textbf{69.6} & 59.9 & \textbf{87.8} & 86.5 & 27.2 & 66.8 & 63.9 \\
        \hline

\textit{Ours} & \textbf{53.5} & 69.2 & 62.2  & 85.7 & 85.4 & 32.5 & \textbf{68.1} & \textbf{64.7}  \\
    \hline
    \hline
    \end{tabular}
    }
    \vspace{-4mm}

    \label{tab:exp_visda}
\end{table}

\subsection{Experimental Details}
\label{sec:experimental_setup}
\vspace{-1mm}

\noindent \textbf{a) Datasets.} \textbf{Office-31} \cite{office} consists of 31 categories of images in three different domains: Amazon (\textbf{A}), Webcam (\textbf{W}) and DSLR (\textbf{D}). \textbf{Office-Home} \cite{venkateswara2017deep} is a more challenging dataset containing 65 classes from four domains: Real World (\textbf{Re}), Art (\textbf{Ar}), Clipart (\textbf{Cl}) and Product (\textbf{Pr}). 
\textbf{VisDA} \cite{visda} comprises of 12 categories of images from two domains: Real (\textbf{R}), Synthetic (\textbf{S}).
The label sets $\mathcal{C}_s$, $\mathcal{C}_t$ are in line with \cite{sta_open_set} and \cite{saito2018open} for all our comparisons. See Suppl. for sample images and further details.

 

\vspace{1mm}

\noindent \textbf{b) Implementation.} We implement the framework in PyTorch and use ResNet-50~\cite{he2016deep_resnet} (till the last pooling layer) as the backbone models $M_s$ and $M_t$ for \textbf{Office-31} and \textbf{Office-Home}, and VGG-16~\cite{vgg} for \textbf{VisDA}. For \textit{inheritable} model training, we use a batch size of $64$ ($32$ source and \textit{negative} instances each), and use the hyperparameters $d=15$ and $K=4|\mathcal{C}_s|$.
During adaptation, we use a batch size of 32 and set the hyperparameter $k=15$. We normalize the instance weights $w(x_t)$ with the $\operatorname{max}$ weight of each batch $B$, \ie $w(x_t)/\max_{x_t \in B}w(x_t)$.
During inference, an \textit{unknown} label is assigned if $\hat{y}_t = \operatorname{argmax}_{c_i}[\sigma(G(F_t(x_t)))]_{c_i}$ is one of the $K$ \textit{negative} classes, otherwise, a shared class label is predicted.
See Supplementary for more details.


\vspace{1mm}
\noindent \textbf{c) Metrics.} 
In line with \cite{saito2018open}, 
we compute the \textit{open-set} accuracy (\textbf{OS}) by averaging the class-wise target accuracy for $|\mathcal{C}_s|+1$ classes (considering target-\textit{unknown} as a single class). Likewise, the shared accuracy (\textbf{OS*}) is computed as the class-wise average of target-shared classes ($\mathcal{C}_t^{sh}=\mathcal{C}_s$).


\subsection{Results}
\vspace{-1mm}
\noindent \textbf{a) State-of-the-art comparison.} In Tables \ref{tab:exp_office31}{\color{red}-}\ref{tab:exp_visda}, we compare against the state-of-the-art UODA method STA~\cite{sta_open_set}. The results for other methods are taken from ~\cite{sta_open_set}. Particularly, in Table~\ref{tab:exp_office31}, we report the mean and std. deviation of \textbf{OS} and \textbf{OS*} over 3 separate runs. Due to space constraints, we report only \textbf{OS} in Table \ref{tab:exp_officehome}. It is evident that adaptation using an \textit{inheritable} model outperforms prior arts that assume access to both \textit{vendor's} data (source domain) and \textit{client's} data (target domain) simultaneously. The superior performance of our method over STA is described as follows. STA learns a domain-agnostic feature extractor by aligning the two domains using an adversarial discriminator. This restricts the model's flexibility to capture the diversity in the target domain, owing to the need to generalize across two domains, on top of the added training difficulties of the adversarial process. In contrast, we employ a target-specific feature extractor ($F_t$) which allows the target predictor to effectively \textit{tune} to the target domain, while \textit{inheriting} the class-separability knowledge. Thus, \textit{inheritable} models offer an effective solution for UODA in practice.

\begin{figure}
    \vspace{-2mm}
    \centering
    \includegraphics[width=\columnwidth]{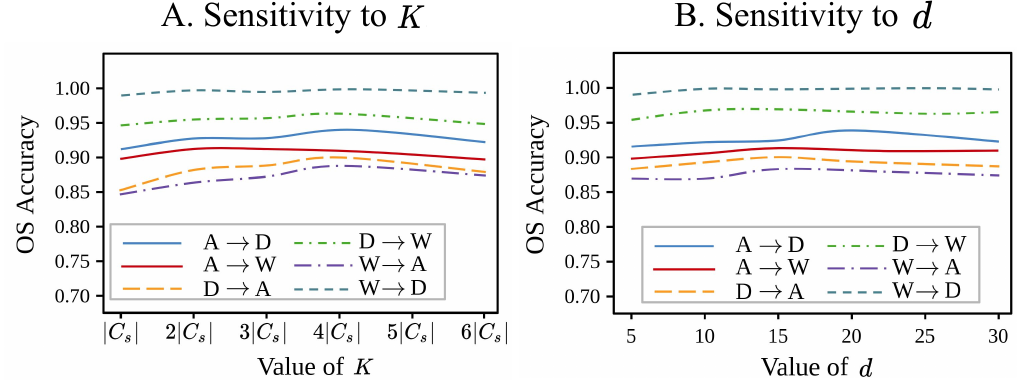}
    \caption{Sensitivity to hyperparam. $K$, $d$. Best viewed in color.}
    \vspace{-3mm}
    \label{fig:exp_Kd_sensitivity}
\end{figure}

\vspace{1mm}
\noindent \textbf{b) Hyperparameter sensitivity.} 
In Fig.~\ref{fig:exp_Kd_sensitivity}, we plot the adaptation performance (\textbf{OS}) on a range of hyperparameter values used to train the \textit{vendor's} model ($K$, $d$). A low sensitivity to these hyperparameters highlights the reliability of the \textit{inheritable} model. In Fig.~\ref{fig:exps_openness_A_sensitivity}{\color{red}C}, we plot the adaptation performance (\textbf{OS}) on a range of values for $k$ on \textbf{Office-31}. Specifically, $k=0$ denotes the ablation where $\mathcal{L}_{inh}$ is not enforced. Clearly, the performance improves on increasing $k$ which corroborates the benefit of \textit{inheriting} class-separability knowledge during adaptation.

\vspace{1mm} 
\noindent \textbf{c) Openness ($\mathbb{O}$).} 
In Fig.~\ref{fig:exps_openness_A_sensitivity}{\color{red}A}, we report the \textbf{OS} accuracy on varying levels of Openness~\cite{scheirer2012toward_openness} $\mathbb{O} = 1 - {|\mathcal{C}_s|} / {|\mathcal{C}_t}|$. 
Our method performs well for a wide range of Openness, owing to the ability to effectively mitigate \textit{negative-transfer}.

\vspace{1mm}
\noindent \textbf{d) Domain discrepancy.}
As discussed in \cite{ben2007analysis}, the empirical domain discrepancy can be approximated using the Proxy $\mathcal{A}$-distance $\hat{d}_{\mathcal{A}}=2(1-2\epsilon)$ where $\epsilon$ is the generalization error of a domain discriminator. 
We compute the \textit{PAD} value at the pre-classifier space for both target-shared and target-\textit{unknown} instances in Fig.~\ref{fig:tsne_vis}{\color{red}B} following the procedure laid out in \cite{ganin2016domain}. The \textit{PAD} value evaluated for target-shared instances using our model is much lower than a source-trained ResNet-50 model, while that for target-\textit{unknown} is higher than a source-trained ResNet-50 model. This suggests that adaptation aligns the source and the target-shared distributions, while separating out the target-\textit{unknown} instances. 


\subsection{Discussion}
\vspace{-1mm}




\begin{figure*}[ht!]
    \centering
    \includegraphics[width=0.95\linewidth]{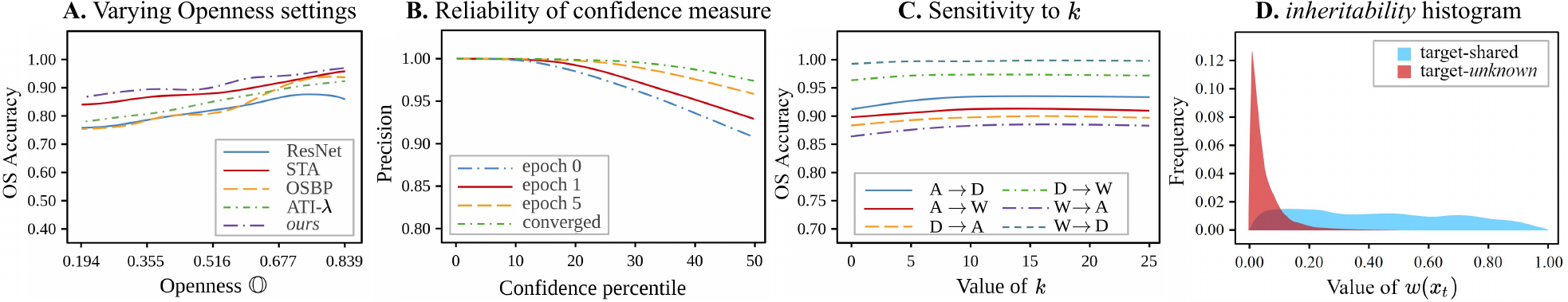}
    \caption{\small \textbf{A)} Adaptation performance on various Openness $\mathbb{O}$ settings for \textbf{Office-31}. \textbf{B)} Precision of the target predictor for the top confidence percentile target instances during adaptation on the \textbf{A}$\rightarrow$\textbf{D} task of \textbf{Office-31}. \textbf{C)} Sensitivity to the hyperparam. $k$. \textbf{D)} Histogram of instance-level \textit{inheritability} values for target-shared and target-\textit{unknown} instances on \textbf{A}$\rightarrow$\textbf{D} task. Best viewed in color.}
    \vspace{-1mm}
    \label{fig:exps_openness_A_sensitivity}
\end{figure*}

\begin{figure*}[t!]
    \centering
    \includegraphics[width=0.95\linewidth]{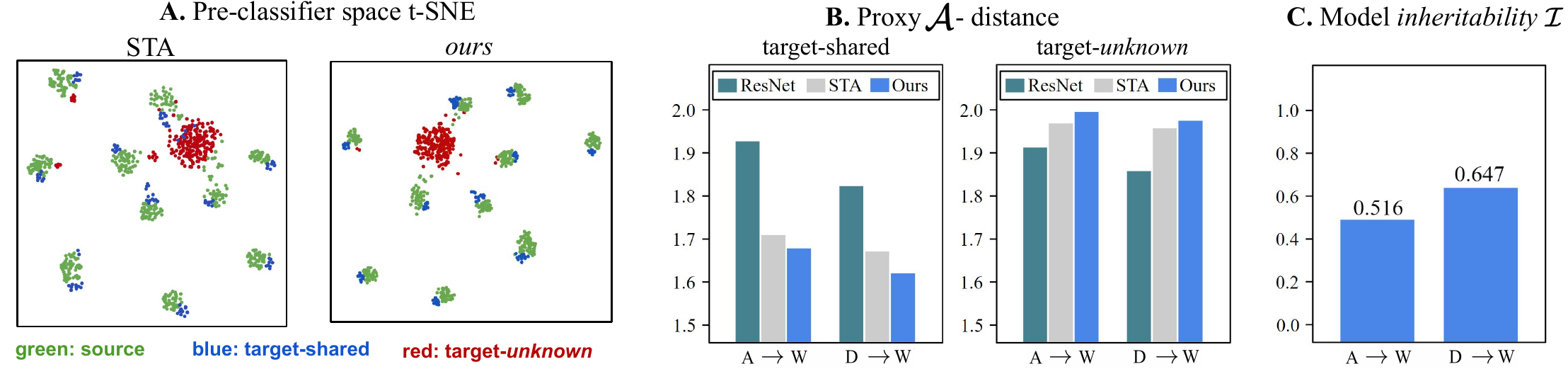}
    \vspace{-1mm}
    \caption{\small Results on \textbf{Office-31}. \textbf{A)} t-SNE~\cite{tsne} visualization of the pre-classifier space for the task \textbf{A}$\rightarrow$\textbf{D} (red: target-\textit{unknown}, blue: target-shared, green: source). \textbf{B)} Proxy $\mathcal{A}$-distance for target-shared (lower is better) and target-\textit{unknown} instances (higher is better) at the pre-classifier space. \textbf{C)} Model \textit{inheritability} of \textit{vendor} models trained on \textbf{A} and \textbf{D} measured for the target \textbf{W}. Best viewed in color.}
    \label{fig:tsne_vis}
    \vspace{-3mm}
\end{figure*}

\noindent \textbf{a) Model inheritability} ($\mathcal{I}$)\textbf{.} Following the intuition in Sec.~\ref{sec:client_adapts}{\color{red}a}, we evaluate the model \textit{inheritability} ($\mathcal{I}$) for the tasks \textbf{D}$\rightarrow$\textbf{W} and \textbf{A}$\rightarrow$\textbf{W} on \textbf{Office-31}. In Fig.~\ref{fig:tsne_vis}{\color{red}C} we observe that for the target \textbf{W}, an \textit{inheritable} model trained on the source \textbf{D} exhibits a higher $\mathcal{I}$ value than that trained on the source \textbf{A}. Consequently, the adaptation task \textbf{D}$\rightarrow$\textbf{W} achieves a better performance than \textbf{A}$\rightarrow$\textbf{W}, suggesting that a \textit{vendor} model with a higher model \textit{inheritability} is a better candidate to perform adaptation to a given target domain.
Thus, given an array of \textit{inheritable} \textit{vendor} models, a \textit{client} can reliably choose the most suitable model for the target domain by measuring $\mathcal{I}$.
The ability to choose a \textit{vendor} model without requiring the \textit{vendor's} source data enables the application of the \textit{vendor-client} paradigm in practice.

\vspace{1mm}
\noindent \textbf{b) Instance-level inheritability} ($w$)\textbf{.} In Fig.~\ref{fig:exps_openness_A_sensitivity}{\color{red}D}, we show the histogram of $w(x_t)$ values plotted separately for target-shared and target-\textit{unknown} instances, for the task \textbf{A}$\rightarrow$\textbf{D} in \textbf{Office-31} dataset. This empirically validates our intuition that the classifier confidence of an \textit{inheritable} model follows the inequality in Eq.~\ref{eq:inheritability_inequality}, at least for the extent of \textit{domain-shift} in the available standard datasets.

\vspace{1mm}
\noindent \textbf{c) Reliability of $w$.} Due to the mitigation of overconfidence issue, we find the classifier confidence to be a good candidate for selecting target sample for pseudo-labeling. 
In Fig.~\ref{fig:exps_openness_A_sensitivity}{\color{red}B}, we plot the prediction accuracy of the top-$k$ percentile target instances based on target predictor confidence ($\max_{c_i\in\mathcal{C}_s}[\sigma(G(F_t(x_t)))]_{c_i}$). Particularly, the plot for epoch-$0$ shows the pseudo-labeling precision, since the target predictor is initialized with the parameters of the source predictor. It can be seen that the top-15 percentile samples are predicted with a precision close to 1.
As adaptation proceeds, $\mathcal{L}_{tune}$ improves the prediction performance of the target model, which can be seen as a rise in the plot in Fig.~\ref{fig:exps_openness_A_sensitivity}{\color{red}B}. Therefore, the bound in Eq.~\ref{result_1} is tightened during adaptation. This verifies our intuition in Sec.~\ref{theoretical_analysis}

\vspace{1mm}
\noindent \textbf{d) Qualitative results.} In Fig.~\ref{fig:tsne_vis}{\color{red}A} we plot the t-SNE~\cite{tsne} embeddings of the last hidden layer (pre-classifier) features of a target predictor trained using STA~\cite{sta_open_set} and our method, on the task \textbf{A}$\rightarrow$\textbf{D}. Clearly, our method performs equally well in spite of the unavailability of source data during adaptation, suggesting that \textit{inheritable} models can indeed facilitate adaptation in the absence of a source dataset.

\vspace{1mm}
\noindent \textbf{e) Training time analysis.} We show the benefit of using \textit{inheritable} models, over a source dataset. Consider a \textit{vendor} with a labeled source domain \textbf{A}, and two \textit{clients} with the target domains \textbf{D} and \textbf{W} respectively. Using the state-of-the-art method STA~\cite{sta_open_set} (which requires labeled source dataset), the time spent by each \textit{client} for adaptation using source data is 575s on an average (1150s in total). In contrast, our method (a single \textit{vendor} model is shared with both the \textit{clients}) results in 250s of \textit{vendor's} source training time (feature-splicing: 77s, $K$-means: 66s, training: 154s), and an average of 69s for adaptation by each client (138s in total). Thus, \textit{inheritable} models provide a much more efficient pipeline by reducing the cost on source training in the case of multiple \textit{clients} (STA: 1150s, ours: 435s). See Supplementary for experiment details.

\section{Conclusion}
\vspace{-1mm}

In this paper we introduced a practical \textit{vendor-client} paradigm, and proposed \textit{inheritable} models to address open-set DA in the absence of co-existing source and target domains. Further, we presented an objective way to measure \textit{inheritability} which enables the selection of a suitable source model for a given target domain without the need to access source data. Through extensive empirical evaluation, we demonstrated state-of-the-art open-set DA performance using \textit{inheritable} models. As a future work, \textit{inheritable} models can be extended to problems involving multiple \textit{vendors} and multiple \textit{clients}.

\vspace{1mm}
\noindent \textbf{Acknowledgements.} This work is supported by a Wipro PhD Fellowship (Jogendra) and a grant from Uchhatar Avishkar Yojana (UAY, IISC\_010), MHRD, Govt. of India.

{\small
\bibliographystyle{ieee_fullname}
\bibliography{egbib}
}



\section*{Supplementary material}

# Supplementary: Towards Inheritable Models for Open-Set Domain Adaptation

This supplementary is organized as follows:

- Sec. 1: Derivation of Result 1
- Sec. 2: Algorithm and implementation details
    - Architecture (Sec. 2.1, Table 1)
    - Inheritable model training (Sec. 2.2, Algo. 1)
    - Feature-splicing algorithm (Algo. 2)
    - Target domain adaptation (Sec. 2.3, Algo. 3)
- Sec. 3: Alternate methods to generate OOD samples
- Sec. 4: Miscellaneous
    - Training time comparison (Sec. 4.1)
    - Dataset description (Sec. 4.2)

## 1. Derivation of Result 1

We derive the result using Theorem 3 in [1], which provides a generalized learning bound for a given pair of source and target domains. We begin by quoting the theorem, following which we apply the theorem to our setting. For quoting the theorem, we use the notations given in [1].

**Theorem 3 [1].** *Let $\mathcal{H}$ be a hypothesis space of VC dimension $d$. Let $\mathcal{U}_S$ and $\mathcal{U}_T$ be unlabeled samples of size $m'$ each, drawn from $\mathcal{D}_S$ and $\mathcal{D}_T$ respectively. Let $S$ be a labeled sample of size $m$ generated by drawing $\beta m$ points from $\mathcal{D}_T$ and $(1-\beta)m$ points from $\mathcal{D}_S$ and labeling them according to $f_S$ and $f_T$, respectively. If $\hat{h} \in \mathcal{H}$ is the empirical minimizer of $\hat{\epsilon}_\alpha(h)$ on $S$ and $h_T^* = \min_{h \in \mathcal{H}} \epsilon_T(h)$ is the target error minimizer, then for any $\delta \in (0,1)$, with probability at least $1 - \delta$ (over the choice of samples),*

$$\epsilon_T(\hat{h}) \leq \epsilon_T(h_T^*) + A + 2(1-\alpha)B \quad (1)$$

where,

$$A = 4\sqrt{\frac{\alpha^2}{\beta} + \frac{(1-\alpha)^2}{1-\beta}} \sqrt{\frac{2d\log(2(m+1)) + 2\log(8/\delta)}{m}} \quad (2)$$

$$B = \frac{1}{2}\hat{d}_{\mathcal{H}\Delta\mathcal{H}}(\mathcal{U}_S, \mathcal{U}_T) + 4\sqrt{\frac{2d\log(2m') + \log(8/\delta)}{m'}} + \lambda \quad (3)$$

and the error $\hat{\epsilon}_\alpha(h) = \alpha\hat{\epsilon}_T(h) + (1-\alpha)\hat{\epsilon}_S(h)$, with $\alpha$ being the relative importance given to the empirical target error. Here, $f_S$ and $f_T$ are the ground-truth labeling functions for the source and the target domains. Further note that, the notations used to describe the errors $\epsilon_T$ and $\hat{\epsilon}_\alpha$ are different from the notations we use in our paper (for instance, we refer to the target error as $\xi_q$). See [1] for more details.

In our formulation, we make two observations. Firstly, the objective of our adaptation step is to improve the performance on the target domain. This entails a choice of $\alpha = 1$ for defining the empirical error. Secondly, we do not have any data from the source domain, which implies $\beta = 1$.

Now, with $\alpha = 1$, the last term in Eq. 1 vanishes. Further, to evaluate Eq. 2 with $\alpha = 1, \beta = 1$ which obtains an indeterminate form, we take the limit as $\alpha \to 1, \beta \to 1$,

$$\lim_{\alpha \to 1, \beta \to 1} A = 4\sqrt{\frac{2d\log(2(m+1)) + 2\log(8/\delta)}{m}} \quad (4)$$

This reduces Eq. 1 to,

$$\epsilon_T(\hat{h}) \leq \epsilon_T(h_T^*) + 4\sqrt{\frac{2d\log(2(m+1)) + 2\log(8/\delta)}{m}} \quad (5)$$

Now, we describe the result. We argue that the knowledge of class-separability (*i.e.* the knowledge of how the classes are distinguished) is *inheritable*, by demonstrating that the *inheritability* criterion holds for shared classes.

During adaptation, we select the top-$k$ percentile target instances based on the value of $w$. In Sec. 5.3c of the paper, we empirically verify that the pseudo-labeling precision for the top-$k$ target instances is close to 1 (see Fig. 5B of the paper for epoch-0 at $k = 15$). Therefore, the pseudo-labeling process can be considered as obtaining target-shared instances with a small noise in the labels. For these instances, by minimizing $\mathcal{L}_{inh}$, we search the hypothesis space $\mathcal{H}$ for the empirical minimizer of $\xi_{q^{sh}}$, *i.e.* $\hat{h}_t = \text{argmin}_{h \in \mathcal{H}} \hat{\xi}_{q^{sh}}(h)$. Thus, considering the correctly pseudo-labeled target instances as the sample $\mathcal{S}$ of size $m$ in Eq. 5, $\hat{h}_t$ as the empirical minimizer of $\xi_{q^{sh}}$, and $h_t^*$ as the optimal hypothesis for $q^{sh}$ we can obtain the relation in Eq. 5 for the target-shared distribution ($q^{sh}$) as,

1

Table 1. Architecture of vendor and client models. 'FC(Inp, Out)' denotes Fully-Connected Layers with 'Inp' input nodes and 'Out' output nodes. BN denotes BatchNorm layer

| Component | Layers |
|---|---|
| $E_s, E_t$ | FC(2048×1024) → ELU → <br> FC(1024×1024) → BN → ELU → <br> FC(1024×256) → ELU → <br> FC(256×256) → BN → ELU |
| $G_s$ | FC(256×$|\mathcal{C}_s|$) |
| $G_n$ | FC(256×$K$) |

$$\xi_{q^{sh}}(\hat{h}_t) \leq \xi_{q^{sh}}(h_t^*) + 4\sqrt{\frac{2d\log(2(m+1)) + 2\log(8/\delta)}{m}} \quad (6)$$

Hence, we obtain Result 1 of the paper, which is stated as,

**Result 1.** *Let $\mathcal{H}$ be a hypothesis class of VC dimension $d$. Let $\mathcal{S}$ be a labeled sample set of $m$ points drawn from $q^{sh}$. If $\hat{h}_t \in \mathcal{H}$ be the empirical minimizer of $\xi_{q^{sh}}$ on $\mathcal{S}$, and $h_t^* = \text{argmin}_{h \in \mathcal{H}} \xi_{q^{sh}}(h)$ be the optimal hypothesis for $q^{sh}$, then for any $\delta \in (0, 1)$, we have with probability of at least $1 - \delta$ (over the choice of samples),*

$$\xi_{q^{sh}}(\hat{h}_t) \leq \xi_{q^{sh}}(h_t^*) + \underbrace{4\sqrt{\frac{2d\log(2(m+1)) + 2\log(8/\delta)}{m}}}_{\text{the term } \epsilon \text{ in Eq. 1 of the paper}} \quad (7)$$

Note that, as we enforce the loss $\mathcal{L}_{inh}$ while learning the target model $h_t$ during adaptation, we prune the hypothesis space while searching for all such hypotheses which satisfy the above condition. This is the *inheritability* criterion obtained using target-shared instances, since we achieve the condition using unlabeled target instances $\mathcal{D}_t$ and the source model $h_s$ (which pseudo-labels the target instances). In this manner, the *inheritability* criterion is satisfied for the target-shared instances, making the knowledge of class-separability *inheritable* for the adaptation task.

Furthermore, $\mathcal{L}_{tune}$ can be seen as a way of self-supervising the target model. We show in Sec. 5c of the paper that the precision of the target model iteratively increases as a result of adaptation (Fig. 5B of the paper, see "epoch-0" through "converged"). Therefore, the self-supervision yields an increasing number of correctly labeled target-shared instances over iterations. This effectively tightens the bound in Eq. 7 (increasing $m$ in Eq. 7 reduces the last term), resulting in a superior adaptation guarantee.

## 2. Algorithm and implementation details

In this section we provide the pseudo-code for the model training and the feature-splicing algorithm and present the implementation details.

**Algorithm 1** Pseudo-Code for inheritable model training

1: **require:** labeled source dataset $\mathcal{D}_s$, parameters $\theta_{M_s}, \theta_{E_s}, \theta_{G_s}, \theta_{G_n}$ of $M_s, E_s, G_s, G_n$ respectively, hyperparameters $K, d$, and no. of *negative* instances $\eta_u$.

**Step 1:** *Pre-training on the source dataset*

2: **while** $\mathcal{L}_b$ has not converged **do**
3: $\quad (X_s, Y_s) \leftarrow$ batch sampled from $\mathcal{D}_s$
4: $\quad \widehat{Y}_s \leftarrow \sigma(G_s(F_s(X_s)))$
5: $\quad$ compute mean $\mathcal{L}_b$ for the batch using $\widehat{Y}_s$ and $Y_s$
6: $\quad$ **update** $\theta_{M_s}, \theta_{E_s}, \theta_{G_s}$ by minimizing $\mathcal{L}_b$ using the Adam optimizer
7: **end while**

**Step 2:** *Training the inheritable model*

8: $\mathcal{D}_n \leftarrow$ FeatureSplicingAlgorithm($\mathcal{D}_s, M_s, K, d, \eta_u$)
9: **while** $\mathcal{L}_s$ has not converged **do**
10: $\quad (X_s, Y_s) \leftarrow$ batch sampled from $\mathcal{D}_s$
11: $\quad (U_n, Y_n) \leftarrow$ batch sampled from $\mathcal{D}_n$
12: $\quad \widehat{Y}_s \leftarrow \sigma(G(F_s(X_s)))$
13: $\quad \widehat{Y}_n \leftarrow \sigma(G(E_s(U_n)))$
14: $\quad$ compute mean $\mathcal{L}_s$ for the batch using $\widehat{Y}_s, \widehat{Y}_n, Y_s, Y_n$
15: $\quad$ **update** $\theta_{E_s}, \theta_{G_s}, \theta_{G_n}$ by minimizing $\mathcal{L}_s$ using the Adam optimizer
16: **end while**

### 2.1. Architecture

For experiments on **Office-31** and **Office-Home**, we choose ResNet-50 upto the last AvgPool layer (2048 dimensions) as the backbone network for $M_s$. For experiments on **VisDA** dataset, we use the backbone as VGG-16 upto the last pooling layer (of shape $7 \times 7 \times 512$), followed by a global AvgPool along each channel to obtain an output of 512 dimensions. Modules $E_s, E_t, G_s, G_n$ are composed of fully connected layers, batch norm layers and non-linearity (ELU) as shown in Table 1.

### 2.2. Vendor trains an inheritable model

The *vendor* has access to an annotated source dataset $\mathcal{D}_s$ using which the *vendor* trains an *inheritable* model $h_s$. See Algorithm 1 for the pseudo-code. During training, we have the following image augmentations: random rotations, flip, color jitter and random crop. We pre-train the parameters of the components $\{M_s, E_s, G_s\}$ on the source domain by minimizing $\mathcal{L}_b$ (L2-L7 in Algo. 1). We then freeze the backbone $M_s$ and perform feature splicing (L8 in Algo. 1). See Algo. 2 for the pseudo-code for the feature splicing operation. We apply feature-splicing at the last layer of $M_s$ to obtain *negative* instances $u_n$. Specifically, L8-L10 in Algo. 2 shows the feature splicing operation ($\phi_d$), where the top-$d$ percentile activations are replaced. In this manner, we

2

**Algorithm 2** FeatureSplicingAlgorithm
1: **require:** labeled source dataset $\mathcal{D}_s$, parameters $\theta_{M_s}$ of $M_s$, hyperparameter $K, d$, no. of *negative* instances $\eta_u$

▷ *Let $|\mathcal{D}|$ denote the cardinality of a set $\mathcal{D}$, $[\cdot]$ denote the indexing operation, $\cdot||\cdot$ denote the append operation and $\mathcal{M}$ denote the output dimensionality of $M_s$ (i.e. $\mathcal{M} = 2048$ for ResNet-50 and $\mathcal{M} = 512$ for VGG-16)*

**Step 1:** *Generating negative instances*
2: Number of dimensions to splice $\eta_d = \mathcal{M} \times d/100$
3: $U_n \leftarrow \{\}$ ▷ *Empty list*
4: **while** $|U_n| \leq \eta_u$ **do**
5: $\quad (x_s^{c_i}, x_s^{c_j}) \leftarrow$ sample 2 instances from $\mathcal{D}_s$ belonging to different classes *i.e.* $c_i, c_j \in \mathcal{C}_s$ where $i \neq j$
6: $\quad u_s^{c_i} \leftarrow M_s(x_s^{c_i})$
7: $\quad u_s^{c_j} \leftarrow M_s(x_s^{c_j})$
8: $\quad idx \leftarrow$ top $\eta_d$ entries in $\text{argsort}(u_s^{c_i})$ ⎫
9: $\quad u_n \leftarrow u_s^{c_i}$ ⎬ $\phi_d(u_s^{c_i}, u_s^{c_j})$
10: $\quad u_n[idx] \leftarrow u_s^{c_j}[idx]$ ⎭
11: $\quad U_n \leftarrow U_n \,||\, u_n$
12: **end while**

**Step 2:** *Assigning labels to negative instances*
13: Perform $K$-means clustering on $U_n$ and assign a unique *negative* class to each cluster
14: $\mathcal{D}_n \leftarrow \{(u_n, y_n) : u_n \in U_n, y_n = $ *negative* class label of $u_n$ obtained from the previous step$\}$

15: **return** $\mathcal{D}_n$

---

**Algorithm 3** Pseudo-Code for target domain adaptation
1: **require:** unlabeled target dataset $\mathcal{D}_t$, parameters $\theta_{F_s}$, $\theta_{F_t}, \theta_{G_s}, \theta_{G_n}$ of $\{F_s, F_t, G_s, G_n\}$, hyperparameter $k$.

▷ *Let $[\cdot]$ denote the indexing operation, and, $\cdot||\cdot$ denote the append operation*

**Step 1:** *Pseudo-labeling process*
2: Number of instances to pseudo-label $\eta_p = |\mathcal{D}_t| \times k/100$
3: $W \leftarrow \{\}$ ▷ *Empty list*
4: $Y_p \leftarrow \{\}$ ▷ *Empty list*
5: $X_p \leftarrow \{\}$ ▷ *Empty list*

6: **for** $x_t$ in $\mathcal{D}_t$ **do**
7: $\quad \hat{y}_p \leftarrow \sigma(G(F_s(x_t)))$
8: $\quad y_p \leftarrow \text{argmax}_{c_i \in \mathcal{C}_s} \hat{y}_p[c_i]$
9: $\quad w \leftarrow \max_{c_i \in \mathcal{C}_s} \hat{y}_p[c_i]$
10: $\quad W \leftarrow W \,||\, w$
11: $\quad Y_p \leftarrow Y_p \,||\, y_p$
12: $\quad X_p \leftarrow X_p \,||\, x_t$
13: **end for**
14: $idx \leftarrow$ top $\eta_p$ entries in $\text{argsort}(W)$
15: $\mathcal{D}_t^p \leftarrow (X_p[idx], Y_p[idx])$

**Step 2:** *Adaptation process*
16: Initialize $\theta_{F_t}$ from $\theta_{F_s}$
17: **while** $\mathcal{L}_{inh} + \mathcal{L}_{tune}$ has not converged **do**

18: $\quad (X_p, Y_p) \leftarrow$ batch sampled from $\mathcal{D}_t^p$.
19: $\quad \widehat{Y}_p \leftarrow \sigma(G(F_t(X_p)))$
20: $\quad$ compute mean $\mathcal{L}_{inh}$ for the batch using $\widehat{Y}_p, Y_p$

21: $\quad X_t \leftarrow$ batch sampled from $\mathcal{D}_t$
22: $\quad$ **for** each $x_t$ in $X_t$ **do**
23: $\quad\quad \hat{s} \leftarrow \Sigma_{c_i \in \mathcal{C}_s} \sigma(G(F_t(x_t)))[c_i]$
24: $\quad\quad w \leftarrow \max_{c_i \in \mathcal{C}_s} \sigma(G(F_s(x_t)))[c_i]$
25: $\quad\quad z_t^{sh} \leftarrow \sigma(G_s(F_t(x_t)))$
26: $\quad\quad z_t^{uk} \leftarrow \sigma(G_n(F_t(x_t)))$
27: $\quad$ **end for**

28: $\quad$ **for** each $x_t$ in $X_t$ **do** ▷ *Normalize weights*
29: $\quad\quad w'(x_t) \leftarrow w(x_t)/(\max_{x_t \in X_t} w(x_t))$
30: $\quad$ **end for**
31: $\quad$ compute mean $\mathcal{L}_{tune}$ using $w', \hat{s}, z_t^{sh}, z_t^{uk}$

32: $\quad$ **update** $\theta_{F_t}$ by minimizing $\mathcal{L}_{inh} + \mathcal{L}_{tune}$ using the Adam optimizer

33: **end while**

---

generate $\eta_u = 20000$ *negative* instances for **Office-31** and $\eta_u = 50000$ for **Office-Home** and **VisDA**. We label these instances by performing a $K$-means clustering on the obtained features (L13-L14 in Algo. 2). Using $\mathcal{D}_s$ and $\mathcal{D}_n$ we train $\{E_s, G_s, G_n\}$ by minimizing $\mathcal{L}_s$ (L9-L16 in Algo. 1).

### 2.3. Client adapts to the target domain

The *client* has access to unlabeled target dataset $\mathcal{D}_t$ and vendor's *inheritable* model $h_s = \{F_s, G\}$, using which the *client* performs adaptation to the target domain. See Algo. 3 for the pseudo-code. We use image augmentations as mentioned in Sec. 2.2. We obtain pseudo-labeled target instances (top-$k$ percentile, based on the instance-level *inheritability* value $w$) into a collection $\mathcal{D}_t^p$ (L2-L15). Thereafter, during adaptation, each batch contains pseudo-labeled target instances (L18) and unlabeled target instances (L21). We normalize the weights obtained for the unlabeled target instances (as mentioned in Sec. 5.1b of the paper), with the maximum weight in a batch (L28-L30). We then train $F_t$ to adapt to the target domain by minimizing $\mathcal{L}_{inh} + \mathcal{L}_{tune}$.

3

## 3. Alternate methods to generate OOD samples

We argue in Sec. 4.1 of the paper that an *inheritable* model for the task of *open-set* DA should have the ability to mitigate the overconfidence issue. We achieve this by training an *out-of-distribution* (OOD) classifier ($G_n$). We explored potential techniques to generate OOD samples, and as mentioned in Sec. 4.1b in the paper, we found that the feature-splicing technique works well in practice for training an *inheritable* model for *open-set* DA. In Table 2, we report the adaptation performance on **Office-31** using *inheritable* models trained with different OOD generation strategies. Here, we discuss the two other strategies we explored.

**a) Linear interpolation between classes.** We randomly choose a pair of source instances corresponding to different classes, and obtain a *negative* feature by linearly interpolating between the features of the two instances as,

$$u_n = \gamma \times u_s^{c_i} + (1 - \gamma) \times u_s^{c_j} \quad (8)$$

where $\gamma \sim \text{Beta}(2,2)$ as proposed in [6], $u_s^{c_i} = M_s(x_s^{c_i})$, $u_s^{c_j} = M_s(x_s^{c_j})$, and $x_s^{c_i}, x_s^{c_j}$ are two instances sampled from different source classes, *i.e.* $c_i, c_j \in \mathcal{C}_s$ ($i \neq j$). This is inspired by techniques such as *mixup* [7, 6] which encourage less confident predictions on the interpolations of latent features. As reported in Table 2, we find that linear interpolation performs worse than feature-splicing. This is because linear interpolation is less effective in producing OOD samples as it yields features from a constrained region between the source classes (as discussed in Sec. 4.1b of the paper, and shown in Fig. 3A of the paper). In contrast, feature-splicing is able to mimic plausible OOD samples through the suppression of *class-specific* traits.

**b) Random suppression of the most active features.** Given a latent-space feature, we randomly scale down the values of top-15 percentile activations as,

$$u_n = \Gamma \odot u_s \quad (9)$$

where $u_s = M_s(x_s)$ and $\Gamma$ is a weight vector containing $\gamma \sim \text{uniform}(0.2, 0.3)$ at the indices corresponding to the top-15 percentile activations, and 1 elsewhere, and $\odot$ denotes element-wise product. In Table 2, we observe that this performs similar to feature-splicing. Essentially, by suppressing the top activations, we obtain a feature which is devoid of the *class-specific* traits (Sec. 4.1b of the paper) thereby resulting in a plausible OOD sample. However, this method of random suppression requires a hyperparameter search to identify an optimal range for the scale of suppression (in this case, 0.2 to 0.3). The feature-splicing technique is meant to avoid this hyperparameter search, by choosing the appropriate replacement for the class-discriminative activations for a given instance.

Table 2. Adaptation performance (OS) of different OOD generation strategies, on **Office-31** (ResNet-50). $|\mathcal{C}_s| = 10$, $|\mathcal{C}_t| = 20$. See Sec. 3 for Interpolation (Sec. 3a) and Suppression (Sec. 3b).

| Method | A→W | A→D | D→W | W→D | D→A | W→A | Avg |
|---|---|---|---|---|---|---|---|
| Interpolation | 83.4 | 86.7 | 95.1 | 96.9 | 79.3 | 76.2 | 86.3 |
| Suppression | 91.0 | 93.3 | 96.4 | 99.6 | 90.2 | 88.2 | 93.1 |
| Feature-splicing | 91.3 | 94.2 | 96.5 | 99.5 | 90.1 | 88.7 | 93.4 |

## 4. Miscellaneous

In this section, we provide the details of the machine and the datasets used for experiments.

### 4.1. Training time comparison

The machine used to run all our experiments has the following hardware specifications. *CPU*: Intel Core i7-7700K, *GPU*: NVIDIA GeForce GTX 1080 with CUDA v8.0.61.

The training time reported in Sec. 5.3e in the paper is obtained on the machine with the above specifications. For a fair evaluation, we compare the training time required for our method and the previous state-of-the-art method STA [2] on identical settings. We use a batch size of 32, with identical optimizers and learning rates for adaptation.

**a) STA.** The STA method adapts to target domain in two steps. The first step trains a multi-binary classifier and a domain discriminator using both source and target samples (16 instances each, in a batch) and the second step involves adaptation to the target domain using both source and target samples (again, 16 instances from each domain). For both **A→D** and **A→W**, on an average, Step 1 took 177s while Step 2 took 398s.

**b) Ours.** For our method, we fixed all hyperparameters to the values as mentioned in Sec. 5.1b of the paper. Overall, our method is much more efficient in the case of multiple *clients*. This is clearly because in our approach, the source training is done only once (by the *vendor*) requiring about 250s. Following this, both the *clients* use the same *vendor* model to adapt to their respective target domains requiring 69s on an average. This is contrast to STA where each client has to train on the *vendor's* source dataset, requiring additional computation during adaptation.

### 4.2. Dataset description

In our experiments, we follow STA [2] to choose the label sets. See Fig. 1 for sample images from each dataset.

The **Office-31** [4] dataset contains 4652 images from 3 domains: Amazon (**A**), DSLR (**D**) and Webcam (**W**). In alphabetical order, the first 10 classes are used as shared classes and, the classes 21-31 are chosen as target-*unknown*.

The **Office-Home** [5] dataset was curated by crawling the web, and thus exhibits a higher *domain-shift* as compared

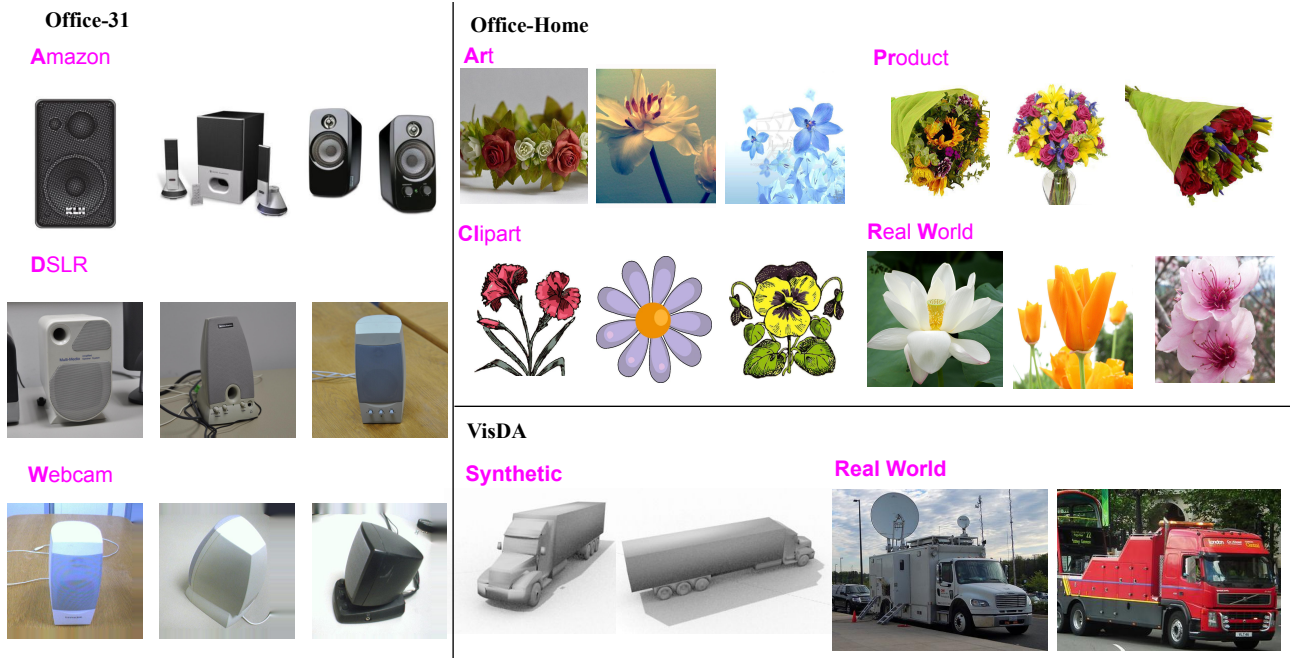

Figure 1. Sample images from the benchmark datasets: **Office-31** showing images belonging to class "Speaker", **Office-Home** showing images belonging to class "Flower" and **VisDA** showing images belonging to class "Truck".

to the **Office-31** dataset. It contains 65 classes of objects with about 15,500 images split into four domains: Art (**Ar**), Clipart (**Cl**), Product (**Pr**) and Real-World (**Rw**). The first 25 classes in alphabetical order are chosen as shared classes and classes 26-65 are chosen as the target-*unknown* classes.

The **VisDA**[3] dataset exhibits a significant amount of *domain-shift* between its two domains: Synthetic (**S**) and Real World (**R**) having about 150k images and 56k images respectively. The Synthetic domain was created by rendering 3D models. The following classes are selected as the shared classes, $\mathcal{C}_s$ = {bicycle, bus, car, motorcycle, train, truck}, while the target-*unknown* classes are chosen as, $\mathcal{C}_t^{uk}$ = {aeroplane, horse, knife, person, plant, skateboard}.

# References

[1] Shai Ben-David, John Blitzer, Koby Crammer, Alex Kulesza, Fernando Pereira, and Jennifer Wortman Vaughan. A theory of learning from different domains. *Machine learning*, 79(1-2):151–175, 2010. 1

[2] Hong Liu, Zhangjie Cao, Mingsheng Long, Jianmin Wang, and Qiang Yang. Separate to adapt: Open set domain adaptation via progressive separation. In *CVPR*, 2019. 4

[3] Xingchao Peng, Ben Usman, Neela Kaushik, Judy Hoffman, Dequan Wang, and Kate Saenko. Visda: The visual domain adaptation challenge. In *CVPRW*, 2018. 5

[4] Kate Saenko, Brian Kulis, Mario Fritz, and Trevor Darrell. Adapting visual category models to new domains. In *ECCV*, 2010. 4

[5] Hemanth Venkateswara, Jose Eusebio, Shayok Chakraborty, and Sethuraman Panchanathan. Deep hashing network for unsupervised domain adaptation. In *CVPR*, 2017. 4

[6] Vikas Verma, Alex Lamb, Christopher Beckham, Amir Najafi, Ioannis Mitliagkas, David Lopez-Paz, and Yoshua Bengio. Manifold mixup: Better representations by interpolating hidden states. In *ICML*, 2019. 4

[7] Hongyi Zhang, Moustapha Cisse, Yann N. Dauphin, and David Lopez-Paz. mixup: Beyond empirical risk minimization. In *ICLR*, 2018. 4

# References

[1] Shai Ben-David, John Blitzer, Koby Crammer, Alex Kulesza, Fernando Pereira, and Jennifer Wortman Vaughan. A theory of learning from different domains. *Machine learning*, 79(1-2):151–175, 2010. 6

[2] Shai Ben-David, John Blitzer, Koby Crammer, and Fernando Pereira. Analysis of representations for domain adaptation. In *NeurIPS*, 2007. 7

[3] Abhijit Bendale and Terrance E Boult. Towards open set deep networks. In *CVPR*, 2016. 7

[4] Fabio Maria Cariucci, Lorenzo Porzi, Barbara Caputo, Elisa Ricci, and Samuel Rota Bulò. Autodial: Automatic domain alignment layers. In *ICCV*, 2017. 2

[5] Woong-Gi Chang, Tackgeun You, Seonguk Seo, Suha Kwak, and Bohyung Han. Domain-specific batch normalization for unsupervised domain adaptation. In *CVPR*, 2019. 1, 2

[6] Hanting Chen, Yunhe Wang, Chang Xu, Zhaohui Yang, Chuanjian Liu, Boxin Shi, Chunjing Xu, Chao Xu, and Qi Tian. Data-free learning of student networks. In *ICCV*, 2019. 4

[7] Boris Chidlovskii, Stéphane Clinchant, and Gabriela Csurka. Domain adaptation in the absence of source domain data. In *ACM SIGKDD*. ACM, 2016. 2

[8] Zhengming Ding and Yun Fu. Deep domain generalization with structured low-rank constraint. *IEEE Transactions on Image Processing*, 27(1):304–313, 2017. 2

[9] Antonio D'Innocente and Barbara Caputo. Domain generalization with domain-specific aggregation modules. In *GCPR*, 2018. 2

[10] Yaroslav Ganin, Evgeniya Ustinova, Hana Ajakan, Pascal Germain, Hugo Larochelle, François Laviolette, Mario Marchand, and Victor Lempitsky. Domain-adversarial training of neural networks. *The Journal of Machine Learning Research*, 17(1):2096–2030, 2016. 2, 7

[11] Yves Grandvalet and Yoshua Bengio. Semi-supervised learning by entropy minimization. In *NeurIPS*, 2005. 5

[12] Kaiming He, Xiangyu Zhang, Shaoqing Ren, and Jian Sun. Deep residual learning for image recognition. In *CVPR*, 2016. 6

[13] Geoffrey Hinton, Oriol Vinyals, and Jeff Dean. Distilling the knowledge in a neural network. *arXiv preprint arXiv:1503.02531*, 2015. 2

[14] Nick Hynes, Raymond Cheng, and Dawn Song. Efficient deep learning on multi-source private data. *arXiv preprint arXiv:1807.06689*, 2018. 2

[15] Lalit P Jain, Walter J Scheirer, and Terrance E Boult. Multiclass open set recognition using probability of inclusion. In *ECCV*, 2014. 2, 7

[16] Aditya Khosla, Tinghui Zhou, Tomasz Malisiewicz, Alexei A Efros, and Antonio Torralba. Undoing the damage of dataset bias. In *ECCV*, 2012. 2

[17] Jakub Konečný, H. Brendan McMahan, Felix X. Yu, Peter Richtarik, Ananda Theertha Suresh, and Dave Bacon. Federated learning: Strategies for improving communication efficiency. In *NeurIPS Workshop on Private Multi-Party Machine Learning*, 2016. 2

[18] Jogendra Nath Kundu, Nishank Lakkakula, and R Venkatesh Babu. Um-adapt: Unsupervised multi-task adaptation using adversarial cross-task distillation. In *ICCV*, 2019. 1

[19] Kimin Lee, Honglak Lee, Kibok Lee, and Jinwoo Shin. Training confidence-calibrated classifiers for detecting out-of-distribution samples. In *ICLR*, 2018. 3, 4

[20] Chun-Liang Li, Wei-Cheng Chang, Yu Cheng, Yiming Yang, and Barnabás Póczos. Mmd gan: Towards deeper understanding of moment matching network. In *NeurIPS*, 2017. 2

[21] Da Li, Yongxin Yang, Yi-Zhe Song, and Timothy M Hospedales. Deeper, broader and artier domain generalization. In *ICCV*, 2017. 2

[22] Da Li, Jianshu Zhang, Yongxin Yang, Cong Liu, Yi-Zhe Song, and Timothy M Hospedales. Episodic training for domain generalization. In *ICCV*, 2019. 2

[23] Yanghao Li, Naiyan Wang, Jianping Shi, Xiaodi Hou, and Jiaying Liu. Adaptive batch normalization for practical domain adaptation. *Pattern Recognition*, 80:109–117, 2018. 2

[24] Hong Liu, Zhangjie Cao, Mingsheng Long, Jianmin Wang, and Qiang Yang. Separate to adapt: Open set domain adaptation via progressive separation. In *CVPR*, 2019. 2, 3, 6, 7, 8

[25] Jonathan Long, Evan Shelhamer, and Trevor Darrell. Fully convolutional networks for semantic segmentation. In *CVPR*, 2015. 1

[26] Mingsheng Long, Yue Cao, Jianmin Wang, and Michael Jordan. Learning transferable features with deep adaptation networks. In *ICML*, 2015. 1

[27] Mingsheng Long, Han Zhu, Jianmin Wang, and Michael I Jordan. Unsupervised domain adaptation with residual transfer networks. In *NeurIPS*, 2016. 2, 5, 7

[28] Mingsheng Long, Han Zhu, Jianmin Wang, and Michael I Jordan. Deep transfer learning with joint adaptation networks. In *ICML*, 2017. 2

[29] Raphael Gontijo Lopes, Stefano Fenu, and Thad Starner. Data-free knowledge distillation for deep neural networks. In *NeurIPS*, 2017. 2

[30] Laurens van der Maaten and Geoffrey Hinton. Visualizing data using t-sne. *Journal of machine learning research*, 9(Nov):2579–2605, 2008. 8

[31] Krikamol Muandet, David Balduzzi, and Bernhard Schölkopf. Domain generalization via invariant feature representation. In *ICML*, 2013. 2

[32] Jogendra Nath Kundu, Phani Krishna Uppala, Anuj Pahuja, and R Venkatesh Babu. Adadepth: Unsupervised content congruent adaptation for depth estimation. In *CVPR*, 2018. 1

[33] Gaurav Kumar Nayak, Konda Reddy Mopuri, Vaisakh Shaj, Venkatesh Babu Radhakrishnan, and Anirban Chakraborty. Zero-shot knowledge distillation in deep networks. In *ICML*, 2019. 2

[34] Sinno Jialin Pan and Qiang Yang. A survey on transfer learning. *IEEE Transactions on knowledge and data engineering*, 22(10):1345–1359, 2009. 2

[35] Pau Panareda Busto and Juergen Gall. Open set domain adaptation. In *ICCV*, 2017. 1, 2, 7

[36] Xingchao Peng, Ben Usman, Neela Kaushik, Judy Hoffman, Dequan Wang, and Kate Saenko. Visda: The visual domain adaptation challenge. In *CVPRW*, 2018. 6

[37] Subhankar Roy, Aliaksandr Siarohin, Enver Sangineto, Samuel Rota Bulo, Nicu Sebe, and Elisa Ricci. Unsupervised domain adaptation using feature-whitening and consensus loss. In *CVPR*, 2019. 2

[38] Kate Saenko, Brian Kulis, Mario Fritz, and Trevor Darrell. Adapting visual category models to new domains. In *ECCV*, 2010. 6

[39] Kuniaki Saito, Shohei Yamamoto, Yoshitaka Ushiku, and Tatsuya Harada. Open set domain adaptation by backpropagation. In *ECCV*, 2018. 1, 2, 3, 6, 7

[40] Swami Sankaranarayanan, Yogesh Balaji, Carlos D. Castillo, and Rama Chellappa. Generate to adapt: Aligning domains using generative adversarial networks. In *CVPR*, June 2018. 2

[41] Walter J Scheirer, Anderson de Rezende Rocha, Archana Sapkota, and Terrance E Boult. Toward open set recognition. *IEEE transactions on pattern analysis and machine intelligence*, 35(7):1757–1772, 2012. 7

[42] Hidetoshi Shimodaira. Improving predictive inference under covariate shift by weighting the log-likelihood function. *Journal of statistical planning and inference*, 90(2):227–244, 2000. 1

[43] Karen Simonyan and Andrew Zisserman. Very deep convolutional networks for large-scale image recognition. In *ICLR*, 2015. 6

[44] Baochen Sun and Kate Saenko. Deep coral: Correlation alignment for deep domain adaptation. In *ECCV*, 2016. 1, 2

[45] A. Torralba and A. A. Efros. Unbiased look at dataset bias. In *CVPR*, 2011. 1

[46] Eric Tzeng, Judy Hoffman, Trevor Darrell, and Kate Saenko. Simultaneous deep transfer across domains and tasks. In *ICCV*, 2015. 2

[47] Eric Tzeng, Judy Hoffman, Kate Saenko, and Trevor Darrell. Adversarial discriminative domain adaptation. In *CVPR*, 2017. 1, 2

[48] Eric Tzeng, Judy Hoffman, Ning Zhang, Kate Saenko, and Trevor Darrell. Deep domain confusion: Maximizing for domain invariance. *arXiv preprint arXiv:1412.3474*, 2014. 1, 2

[49] Hemanth Venkateswara, Jose Eusebio, Shayok Chakraborty, and Sethuraman Panchanathan. Deep hashing network for unsupervised domain adaptation. In *CVPR*, 2017. 6

[50] Vikas Verma, Alex Lamb, Christopher Beckham, Amir Najafi, Ioannis Mitliagkas, David Lopez-Paz, and Yoshua Bengio. Manifold mixup: Better representations by interpolating hidden states. In *ICML*, 2019. 4

[51] Hongliang Yan, Yukang Ding, Peihua Li, Qilong Wang, Yong Xu, and Wangmeng Zuo. Mind the class weight bias: Weighted maximum mean discrepancy for unsupervised domain adaptation. In *CVPR*, 2017. 2

[52] Jason Yosinski, Jeff Clune, Yoshua Bengio, and Hod Lipson. How transferable are features in deep neural networks? In *NeurIPS*. 2014. 1

[53] Kaichao You, Mingsheng Long, Zhangjie Cao, Jianmin Wang, and Michael I. Jordan. Universal domain adaptation. In *CVPR*, June 2019. 2, 3

[54] Matthew D Zeiler and Rob Fergus. Visualizing and understanding convolutional networks. In *ECCV*, 2014. 4

[55] Hongyi Zhang, Moustapha Cisse, Yann N. Dauphin, and David Lopez-Paz. mixup: Beyond empirical risk minimization. In *ICLR*, 2018. 4

[56] Quanshi Zhang, Ying Nian Wu, and Song-Chun Zhu. Interpretable convolutional neural networks. In *CVPR*, 2018. 4



\end{document}